\documentclass[10pt,journal,twoside,final]{IEEEtran}
\usepackage[linesnumbered,ruled,vlined]{algorithm2e}
\usepackage[cmex10]{amsmath}
\usepackage[subnum]{cases}

\usepackage{cite,amsfonts,array,amssymb,soul,
algpseudocode,color,multirow,url, subfig,enumitem,bigstrut,graphicx,booktabs,lipsum,float}

\newcolumntype{C}[1]{>{\centering\arraybackslash}p{#1}}

\usepackage{colortbl}
\usepackage{nameref}

\AtBeginDocument{
  \providecommand\BibTeX{
    {
      \normalfont B\kern-0.5em{\scshape i\kern-0.25em b}\kern-0.8em\TeX
    }
  }
}
\graphicspath{{figures/}}
\SetKwInput{KwInput}{Input} %
\SetKwInput{KwOutput}{Output}
\newcolumntype{L}[1]{>{\raggedright\let\newline\\\arraybackslash\hspace{0pt}}m{#1}}
\newcolumntype{R}[1]{>{\raggedleft\let\newline\\\arraybackslash\hspace{0pt}}m{#1}}

\begin{document}

\title{Practical Wi-Fi-based Motion Recognition Under Variable Traffic Patterns}

\author{Guolin~Yin,~\IEEEmembership{Member,~IEEE},
  Junqing~Zhang,~\IEEEmembership{Senior~Member,~IEEE}, Guanxiong~Shen,~\IEEEmembership{Member,~IEEE}, and Simon~L.~Cotton,~\IEEEmembership{Fellow,~IEEE}
  \thanks{Manuscript received xxx; revised xxx; accepted xxx. Date of publication xxx; date of current version xxx.
    The work of J. Zhang was supported in part by the UK Engineering and Physical Sciences Research Council (EPSRC) under grant ID EP/V027697/1 and EP/Y037197/1.
    The work was completed when Guolin Yin was pursuing his PhD at the University of Liverpool. The review of this paper was coordinated by xxx.
  (\textit{Corresponding author: Junqing Zhang})}
  \thanks{G.~Yin and S.~L.~Cotton are with the Centre for Wireless Innovation (CWI), School of Electronics, Electrical Engineering and Computer Science, Queen’s University Belfast, Belfast, BT3 9DT, U.K. (e-mail: \{Guolin.Yin, simon.cotton\}@qub.ac.uk)}
  \thanks{J.~Zhang is with the School of Computer Science and Informatics, University of Liverpool, Liverpool, L69 3DR, United Kingdom. (email: junqing.zhang@liverpool.ac.uk)}
  \thanks{G.~Shen is with the School of Cyber Science and Engineering, Southeast University, Nanjing, 210096, China.  (e-mail: gxshen@seu.edu.cn)}
  \thanks{Color versions of one or more of the figures in this paper are available online at http://ieeexplore.ieee.org.}
  \thanks{Digital Object Identifier xxx}
}

\maketitle

\begin{abstract}
  Wi-Fi sensing detects human motions and activities by analysing the channel state information (CSI) derived from Wi-Fi transmissions. However, the impact of variable transmission traffic, which determines when CSI can be measured, is often overlooked.
  Existing Wi-Fi sensing systems are trained with fixed input size and sampling rate, which suffer from poor sampling rate generalisation.
  This paper proposes a novel Wi-Fi sensing approach for motion recognition applications, e.g., gesture and activity recognition, under variable traffic patterns. A sampling rate versatile neural network (SRV-NN) based on the transformer is proposed to efficiently handle variable input-sized sensing signals. A dynamic sampling rate augmentation is employed for variable sampling rates and intervals.
  To validate our approach, we have carried out extensive experimental evaluation, using two self-collected datasets, namely SRV Activity and SRV Gesture, as well as two publicly available datasets.
  Our method demonstrated exceptional performance and stability under variable sampling rates, with substantial improvements in average accuracy compared to baseline models without augmentation. The proposed approach significantly enhances stability by greatly reducing accuracy variance across different sampling rates.
\end{abstract}


\section{Introduction}

Motion recognition is crucial for numerous human-centric applications, such as virtual reality, smart homes, and healthcare. Traditional recognition methods such as vision-based~\cite{gkioxari2018detecting} and wearable device-based approaches~\cite{BodyScope}, effectively recognise human activities. Nevertheless, methods based on vision pose risks of privacy violations and have limited sensing ranges, while wearable devices can be inconvenient due to the necessity of wearing sensors.

Recently, Wi-Fi sensing has emerged as a non-intrusive and privacy-preserving alternative that reuses widely deployed Wi-Fi devices without extra sensing hardware~\cite{hernandez2022wifi}. These advantages make it desirable for applications such as human activity recognition in smart homes~\cite{li2021two,wang2018spatial,meneghello2022sharp} and gesture recognition in human-computer interaction (HCI)~\cite{li2020wihf,gao2021towards,signfi,widar}.

\subsection{Problem Statement}
The IEEE 802.11 protocol (Wi-Fi) is inherently designed for data transmission rather than sensing tasks. Therefore, existing Wi-Fi sensing research often relies on packet traffic, specifically created for sensing tasks. However, such sensing traffic will affect the Wi-Fi data communications.

The IEEE 802.11bf amendment has standardised WLAN sensing to achieve integrated sensing and communications (ISAC)~\cite{ieee80211bf2025,du2023overview}. This will enable Wi-Fi networks to merge sensing capabilities with routine data transmission by leveraging existing communication traffic.
This brings a paradigm shift from sensing with dedicated traffic to sensing with application-dependent communication traffic. For instance, streaming 4K video demands many packets, whereas web browsing typically requires a much lower packet transmission rate. However, existing studies often assume a steady and high transmission rate, which does not hold in real-world scenarios; even in laboratory settings, packet loss and CSMA/CA prevent fixed sampling intervals.
Such traffic variations raise three challenges for Wi-Fi-based motion recognition: (1) low sampling rate, (2) sampling rate generalisation, and (3) the fixed-length input of existing models. These challenges are summarised below and analysed in detail in Section~\ref{sec:motivation}.

In this paper, the traffic pattern refers to the packet transmission rate and packet interval. Because we are focusing on the sensing, we also refer to the transmission rate as the sampling rate. Such terminology is used in~\cite{wang2015understanding,widar,8067693} as well. The time interval between adjacent packets is denoted as the sampling interval.

\textit{Low sampling rate.} Intuitively, a high sampling rate leads to good sensing performance, and prior systems adopt rates as high as 2500~Hz and 1000~Hz for human activity recognition~\cite{wang2015understanding,8067693} and 1000~Hz for gesture recognition~\cite{widar,CrossSense}. Such traffic dedicated to sensing, however, consumes valuable bandwidth~\cite{zheng2024pushing,he2023sencom}, leading to network congestion and elevated interference that degrade data transmission quality. The straightforward remedy is to reduce the sampling rate, but the recent work WiGesture~\cite{gao2021towards} reports that sensing performance decreases significantly once the rate drops below 100~Hz, and the same effect has been observed in Widar~\cite{widar}. This is because a lower sampling rate leaves less information in the sensing signal, which causes ambiguity in the sensing task.

\textit{Sampling rate generalisation.} Packet loss and CSMA/CA make the received rate vary, so existing systems upsample the signal to a fixed rate~\cite{falldefi,zheng2016smokey}. This is effective only when the two rates are close. Under ISAC, where sensing reuses ordinary data transmissions, the rate can fall from 1000~Hz to as low as 10~Hz~\cite{falldefi}, a gap that no upsampling can bridge; models trained at a single rate therefore generalise poorly.

\textit{Fixed-length input.} Rate variation changes the number of sampling points in the input. CNN-based sensing models are constrained to fixed-length inputs~\cite{WiDFF,yang2022autofi}, so existing methods resample to a fixed dimension. RNN-based models can inherently process variable-length sequences, but to the best of the authors' knowledge, this capability has not been exploited in Wi-Fi sensing, leaving the variable-length challenge largely unexplored.

\subsection{Contributions}
To address the above-mentioned research challenges, we have designed a sampling rate versatile neural network (SRV-NN). This model is specifically designed to effectively manage variable input lengths, making it highly suitable for scenarios with sampling rate variations. In addition, we have created a data augmentation method, named dynamic sampling rate augmentation, to improve the sampling rate generalisation capability by enriching sampling rate/interval diversity of the training dataset.
We take gesture recognition and human activity recognition as case studies and carry out comprehensive experimental evaluation using two self-collected datasets and two public datasets, i.e., SHARP~\cite{meneghello2022sharp} and Widar~\cite{widar}.
The contributions are outlined as follows.
\begin{itemize}
  \item A transformer-based neural network, SRV-NN, was designed to address the challenge of variable input length caused by sampling rate variations in real-world Wi-Fi traffic. SRV-NN exploits a transformer architecture as a feature extractor, to capture temporal dependencies in CSI sequences of arbitrary lengths, paired with a novel sampling rate versatile classifier that extracts length-invariant features from variable-sized inputs. This design enables robust generalisation across diverse sampling rates under dynamic network conditions.
  \item A Dynamic Sampling Rate Augmentation is designed to tackle the challenge of varying sampling rates by incorporating adaptive sampling rate generation and a stochastic sampling approach. These techniques enhance dataset diversity and improve performance under low and irregular sampling rates. The adaptive sampling rate generation algorithm introduces variation in sampling rates within each training batch, which enriches dataset diversity. Stochastic sampling, which involves randomly selecting sampling points, can significantly improve model performance in situations where sampling rates are low or random. This approach can enhance the system's robustness against sampling rate variations.
  \item Comprehensive experimental validation is provided using two self-collected datasets (SRV Activity and SRV Gesture) and two public datasets (SHARP and Widar). Our results demonstrate that the SRV-NN outperforms traditional deep learning architectures (CNN, BiLSTM, GRU, CNN-GRU) even without augmentation. With augmentation, it achieves high accuracy with substantially reduced variance across diverse sampling rates, confirming robust generalisation under variable traffic patterns.
\end{itemize}

The rest of this paper presents related work (Section~\ref{sec:relatedwork}), motivations (Section~\ref{sec:motivation}), system overview (Section~\ref{sec:overview}), signal preprocessing (Section~\ref{sec:signalprocess}), SRV-NN design (Section~\ref{sec:SRV-NN}), dynamic sampling rate augmentation (Section~\ref{sec:aug}), experimental evaluation (Section~\ref{sec:experiments}), and conclusions (Section~\ref{sec:conclusion}).

  \section{Related Work}\label{sec:relatedwork}
  With the evolution of wireless sensing technology, Wi-Fi-based sensing attracted significant research attention due to the prevalence of Wi-Fi technology.
  Wi-Fi sensing encompasses a broad range of applications, including large-scale movements like human activity recognition~\cite{wang2014eyes}, fall detection~\cite{zhang2019commercial}, and ~{person identification based on posture~\cite{WiDFF}} and gait~\cite{wang2016gait} as well as small-scale motions like gesture recognition~\cite{widar, yin2022fewsense} and sign language recognition~\cite{signfi}. Gesture recognition has emerged as an important application in Wi-Fi sensing~\cite{ma2019wifi}.

  \subsection{Deep Learning-Based Wi-Fi Sensing}\label{subsec:dl_wifi}
  Owing to the robust feature extraction capabilities of Deep Neural Networks (DNNs), they have been extensively utilised in the field of Wi-Fi sensing~\cite{ma2019wifi} as feature extractors to discern fine-grained features from CSI streams. 
  Various architectures have been employed, which can be classified into three categories: spatial models, such as Convolutional Neural Networks (CNNs)~\cite{yin2022fewsense,signfi,9235578}; sequential models based on Recurrent Neural Networks (RNNs), including Long Short-Term Memory (LSTM)~\cite{chen2018wifi, 8067693} and Gated Recurrent Units (GRUs)~\cite{yang2023sensefi}; and hybrid models, such as CNN-GRU~\cite{widar, li2020wihf}. Typically, the extracted features are subsequently classified using fully connected layers~\cite{li2023caring}.

  Researchers have extensively investigated diverse challenges in Wi-Fi sensing. For instance, the authors in \cite{zheng2025segall} studied the segmentation problem, and authors in \cite{lu2024imperceptible,yin2025evasion} examined the security vulnerabilities of Wi-Fi sensing systems. Hu et al. \cite{hu2023muse} exploited near-field Wi-Fi channel variations to enable multi-person sensing.
\label{para:r1c3-domain-variation}
The issue of domain variation has attracted significant attention. This stems from the inherent sensitivity of wireless signals to environmental and contextual changes
~\cite{li2020wihf,CrossSense,kang2021context},
including variations in multipath propagation, subjects, and locations. Such variations can cause substantial changes in signal representation, which in turn degrades the performance of sensing systems. To mitigate this problem, various approaches have been proposed. The Widar system
~\cite{widar}
introduced a domain-independent feature representation based on body-coordinate velocity profiles (BVP) of gestures. Zhang et al.
~\cite{zhangLocationindependentHumanActivity2023}
designed CSI-MTGN, a graph neural network that improves location-independent human activity recognition by addressing location-dependent features. Zheng et al.
~\cite{zheng2024adawifi}
developed AdaWiFi, which enhances cross-environment adaptability in Wi-Fi sensing. Additionally, domain-adversarial training has been employed to extract domain-invariant features
~\cite{jiang2018towards,kang2021context}.

  \subsection[General Time-Series Learning and Data Augmentation]{General Time-Series Learning and Data Augmentation}\label{subsec:general_timeseries}
General time-series learning builds models for sequences of measurements ordered in time, such as wearable-sensor and speech signals. Such models include recurrent and convolutional networks~\cite{iwana2021empirical} and transformer encoders, which learn representations of multivariate time series from the whole sequence~\cite{zerveas2021transformer}. Temporal pooling summarises the features along the time axis into a fixed-size representation for classification~\cite{lee2021learnable}.
\label{para:r1c1-general-prior}

Data augmentation increases the diversity of the training set by adding modified copies of the training series~\cite{iwana2021empirical}. Existing augmentations fall into two groups, according to what they modify. Value-based augmentations change the recorded values: jittering adds random noise~\cite{bishop1995training,an1996effects}, and magnitude scaling multiplies the series by a random factor~\cite{um2017data}. Time-based augmentations change the number or the temporal distribution of the sampling points. Time warping stretches or compresses parts of the series while keeping the number of points~\cite{um2017data}. Resampling changes the sampling rate of the whole series. Window slicing cuts a shorter contiguous segment from the series for training~\cite{leguennec2016data}. These augmentations are not suitable for CSI-based Wi-Fi sensing, because none of them produces a varying sampling rate together with irregular sampling intervals, which is what Wi-Fi traffic causes in CSI. Value-based augmentations keep every sampling point, whereas Wi-Fi traffic determines how many sampling points are captured. Time warping changes the sampling intervals but keeps the number of sampling points. Resampling changes the sampling rate but keeps the intervals equal. Window slicing keeps both the sampling rate and the intervals. Wi-Fi traffic, in contrast, changes the sampling rate and makes the intervals irregular.
\label{para:r1c1-augmentation-distinction}

  \subsection[ISAC in Wi-Fi Networks under Variable Traffic]{ISAC in Wi-Fi Networks under Variable Traffic}\label{subsec:wifi_isac}
ISAC has recently received increasing attention in Wi-Fi networks because the IEEE 802.11bf amendment standardises WLAN sensing and allows sensing functions to coexist with ordinary data communication~\cite{ieee80211bf2025,du2023overview}.
Existing studies on the coexistence of sensing and communication in Wi-Fi networks have explored practical deployment~\cite{he2023sencom} and sensing with low transmission rates~\cite{zheng2024pushing}.
These studies indicate that practical WLAN sensing should avoid dedicated high-rate probing traffic and operate on the traffic of ordinary network services. In that case, the CSI sampling rate and sampling interval depend on communication demand, MAC contention, packet loss and the applications in use.

Several Wi-Fi sensing systems handle sparse or irregular CSI by restoring a regular sequence before recognition. SenCom~\cite{he2023sencom} fits splines to the unevenly spaced CSI of ordinary traffic and resamples it to a fixed rate, and it injects probing packets when the traffic is too sparse for that rate. WiImg2.0~\cite{zheng2024pushing} recovers high-rate CSI from evenly spaced low-rate CSI with a generative adversarial network, and works at rates from 10 to 120~Hz. FallDeFi~\cite{falldefi} and Smokey~\cite{zheng2016smokey} transmit at a fixed rate and interpolate the CSI over lost packets; FallDeFi keeps its accuracy only down to 600~Hz. MUSE-Fi~\cite{hu2023muse} resamples the CSI of bursty traffic onto a regular grid and fills the missing points by sparse recovery, but very sparse traffic may be beyond recovery.
\label{para:r1c1-wifi-prior}

Table~\ref{tab:related_work_gap} compares these systems with this work in terms of irregular CSI input and sampling rate generalisation, i.e., keeping the recognition accuracy across sampling rates. Unlike the systems that restore a regular sequence, this work trains one model to recognise motions directly from CSI at different sampling rates and with irregular sampling intervals.

\begin{table}[!t]
  \centering
  \caption{Comparison with existing Wi-Fi sensing systems that use low-rate or irregular CSI.}
  \label{tab:related_work_gap}
  \begingroup\scriptsize
  \setlength{\tabcolsep}{3pt}
  \resizebox{\columnwidth}{!}{%
    \begin{tabular}{lc>{\centering\arraybackslash}p{1.6cm}>{\centering\arraybackslash}p{2.4cm}}
  \toprule
  \textbf{System} & \textbf{Low-rate or irregular CSI handling} & \textbf{Irregular}\par\textbf{CSI input} & \textbf{Sampling rate}\par\textbf{generalisation} \\
  \midrule
  SenCom~\cite{he2023sencom} & Spline resampling, packet injection & $\checkmark$ & $\times$ \\
  WiImg2.0~\cite{zheng2024pushing} & GAN recovery & $\times$ & $\checkmark$ \\
  FallDeFi~\cite{falldefi} & Linear interpolation of lost packets & $\times$ & $\times$ \\
  Smokey~\cite{zheng2016smokey} & Linear interpolation of lost packets & $\times$ & $\times$ \\
  MUSE-Fi~\cite{hu2023muse} & Resampling, sparse recovery & $\checkmark$ & $\times$ \\
  \midrule
  \textbf{This work} & Sampling rate versatile classifier & $\checkmark$ & $\checkmark$ \\
  \bottomrule
\end{tabular}%
  }
  \endgroup
\end{table}

\section{Background and Motivations}
  \subsection{Background}\label{sec:basic}
  \subsubsection{CSI in Wi-Fi Sensing}
  In the context of Wi-Fi motion recognition, human body movements influence the Wi-Fi signal propagation channel, which will cause variation to the channel frequency response $H(f,t)$. Orthogonal frequency-division multiplexing (OFDM) is used in IEEE 802.11a/g/n/ac/ax.
  Typically, Wi-Fi OFDM operates across bandwidths ranging from 20 MHz to 320 MHz, which are subdivided into $C$ uniformly distributed subcarriers.
  As motions are performed, a Wi-Fi transmitter sends a continuous stream of packets to a receiver, which captures a sequence of CSI data over time, reflecting the motion-induced channel variations. The CSI sequence of one motion forms a sample, given as
  \begin{equation}
    \mathbf{x} =
    \begin{bmatrix}
      H(f_{1}, t_1) & H(f_{2}, t_1) & \cdots & H(f_{C}, t_1) \\
      H(f_{1}, t_2) & H(f_{2}, t_2) & \cdots & H(f_{C}, t_2) \\
      \vdots          &          \vdots & \ddots & \vdots          \\
      H(f_{1}, t_N) & H(f_{2}, t_N) & \cdots & H(f_{C}, t_N) \\
    \end{bmatrix},
  \end{equation}
  where $N$ is the total number of sampling points. The sampling rate $R$ can be calculated as
  \begin{equation}\label{eq:sr}
    R = \frac{N}{T},
  \end{equation}
  where $T$ is the time duration for capturing $\mathbf{x}$.

  The sample $\mathbf{x}$ encapsulates motion-related information that is distinct to each specific motion, with different motions generating unique patterns that affect the CSI in varied ways.
  CNN and RNN have been widely adopted to learn the relationship between motions and corresponding CSI variations due to their powerful feature extraction capabilities, which can classify a wide array of human activities and gestures based on Wi-Fi CSI measurements.

  \subsubsection{DNN-Powered Wi-Fi Sensing}
  \begin{figure}[!t]
    \centering
    \includegraphics[width=0.85\linewidth]{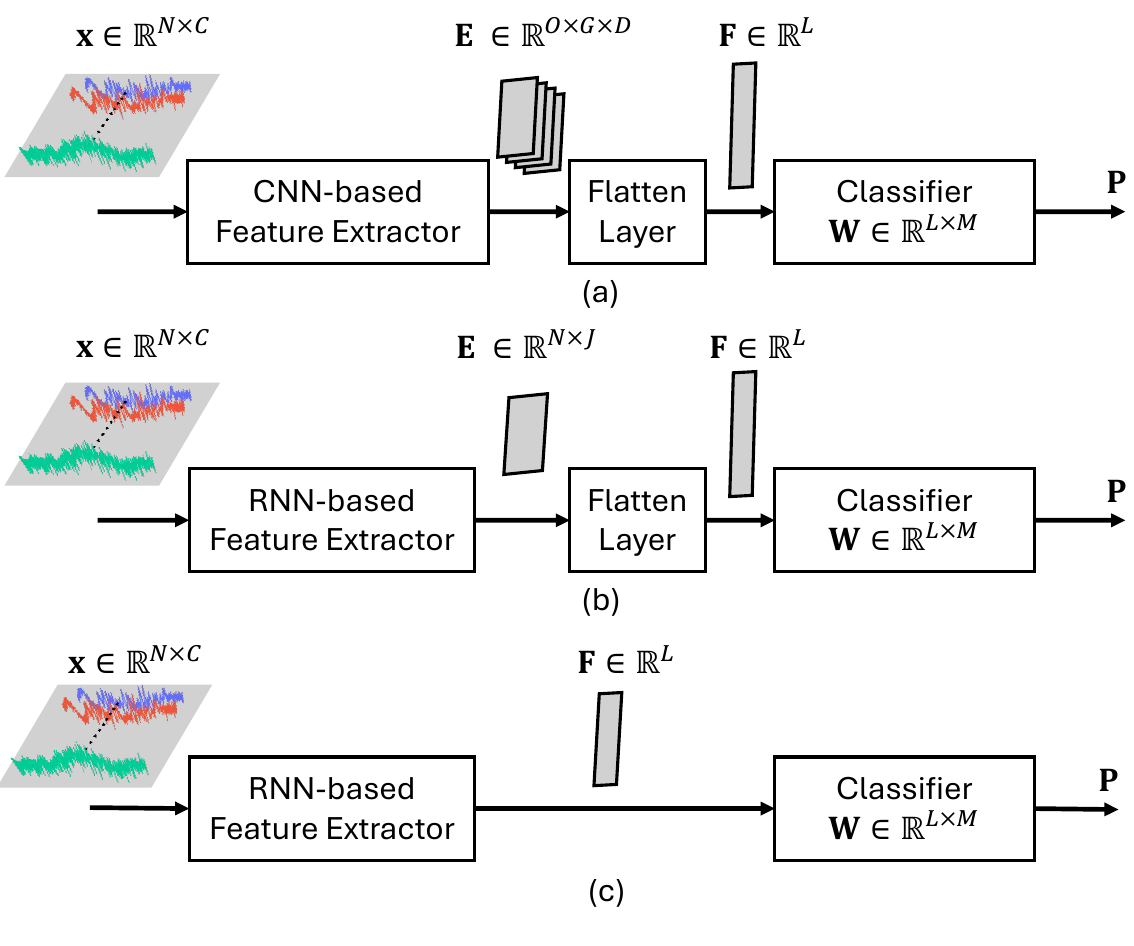}
    \caption{(a) CNN-based Wi-Fi sensing model. (b) RNN-based Wi-Fi sensing model using all hidden states for classification. (c) RNN-based Wi-Fi sensing model using the last hidden state for classification.}
    \label{fig:model}
  \end{figure}

  As illustrated in Fig.~\ref{fig:model}, deep learning-based Wi-Fi sensing models typically consist of a feature extractor followed by a classifier. The feature extractor (CNN or RNN) processes the input CSI data, while the classifier categorises the extracted features into target classes. We describe the two predominant architectures below.

  \textbf{CNN-based models}, i.e., Fig.~\ref{fig:model}(a), process input data $\mathbf{x} \in \mathbb{R}^{N \times C}$ through convolutional layers to generate a feature matrix $\mathbf{E} \in \mathbb{R}^{O \times G \times D}$, where $O$ and $G$ represent the spatial dimensions and $D$ is the number of kernels. This feature matrix is flattened to $\mathbf{F} \in \mathbb{R}^{L}$ where $L = O \times G \times D$. This architecture is widely employed in Wi-Fi sensing applications~\cite{signfi,yin2022fewsense}.

\textbf{RNN-based models}, i.e., Figs.~\ref{fig:model}(b) and (c), process sequential input $\mathbf{x}$ to generate hidden states $\mathbf{E} \in \mathbb{R}^{N \times J}$, where $J$ equals the number of hidden units. Two classification approaches are commonly used:
\begin{itemize}
 \item All hidden states are flattened into $\mathbf{F} \in \mathbb{R}^{L}$ with $L = N \times J$, which captures comprehensive temporal dynamics~\cite{chen2018wifi}, as illustrated in Fig.~\ref{fig:model}(b).
 \item Only the final hidden state is used as $\mathbf{F} \in \mathbb{R}^{L}$ with $L = J$, which focuses on the most recent sequential information~\cite{yang2023sensefi}, as illustrated in Fig.~\ref{fig:model}(c).
\end{itemize}

For all the models, the classifier employs fully connected layers (here simplified as a single layer with weight matrix $\mathbf{W} \in \mathbb{R}^{L \times M}$ and bias $\mathbf{B} \in \mathbb{R}^{M}$) to compute class probabilities:

  \begin{equation}\label{eqn:classification}
    \mathbf{P} = \text{softmax}(\mathbf{W}^{\top} \cdot \mathbf{F} + \mathbf{B}),
  \end{equation}

  where $\mathbf{P} \in \mathbb{R}^{M}$ represents the probability distribution over $M$ target classes. The dimensionality of feature vector $\mathbf{F}$ must align with the first dimension of $\mathbf{W}$ for compatibility.

\subsection{Motivations}\label{sec:motivation}

With the advent of ISAC, sensing and communication occur simultaneously, i.e., share the same Wi-Fi traffic.
However, the existing Wi-Fi sensing requires a constant sampling rate and interval, which faces significant challenges in deployment, as the transmission rate in real-world scenarios is often unpredictable. These fluctuations in the packet transmission rates directly impact the sampling rate $R$, which, in turn, results in changes to the number of sampling points $N$ for motions of the same duration.
\begin{figure}[!t]
  \centering
  \subfloat[]{
    \includegraphics[width=.5\linewidth]{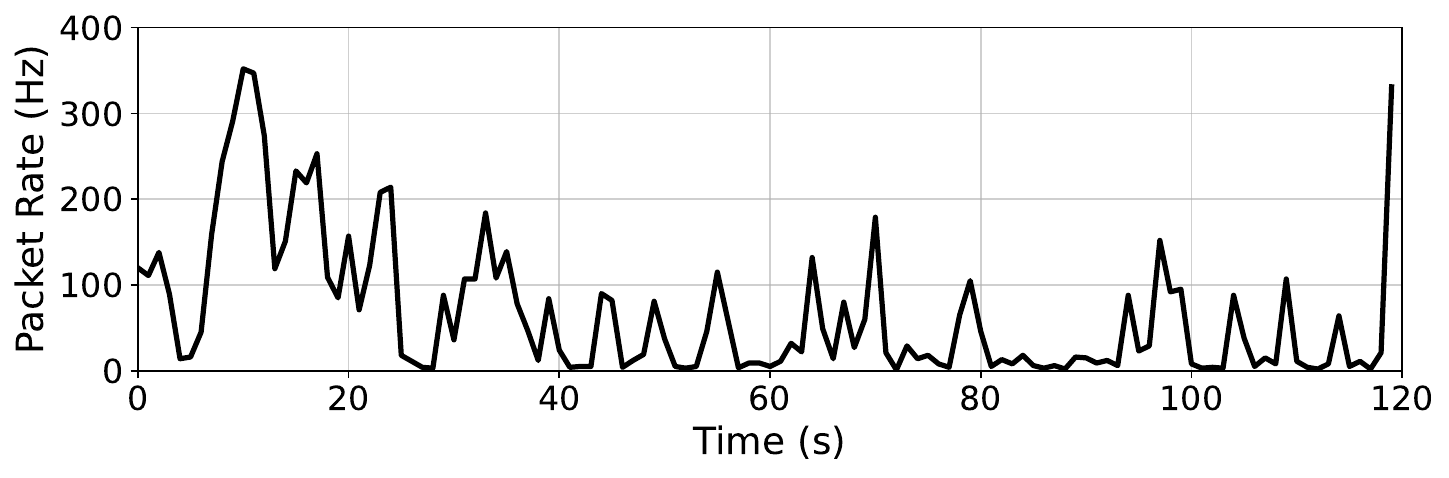}

  }
  \subfloat[]{
    \includegraphics[width=.5\linewidth]{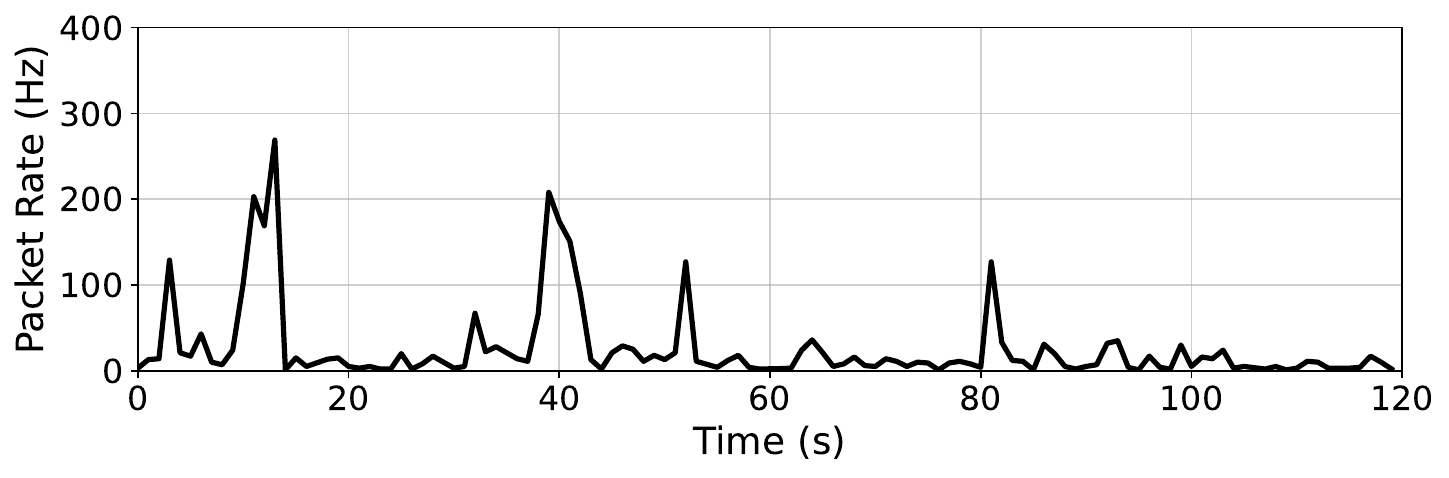}

  }

  \subfloat[]{
    \includegraphics[width=.5\linewidth]{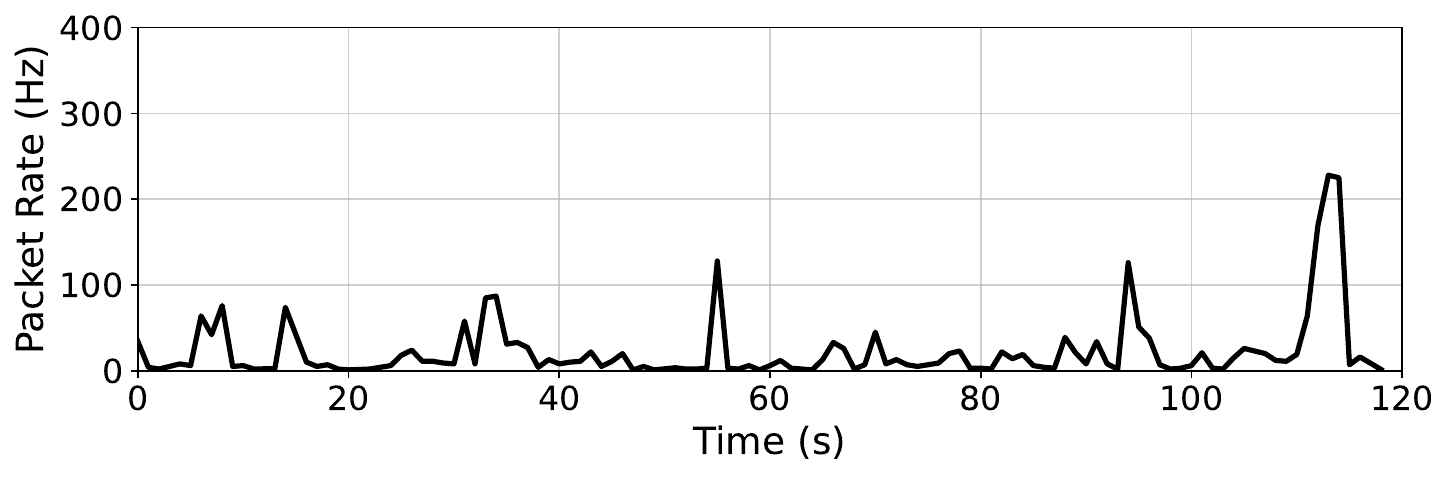}

  }
  \subfloat[]{
    \includegraphics[width=.5\linewidth]{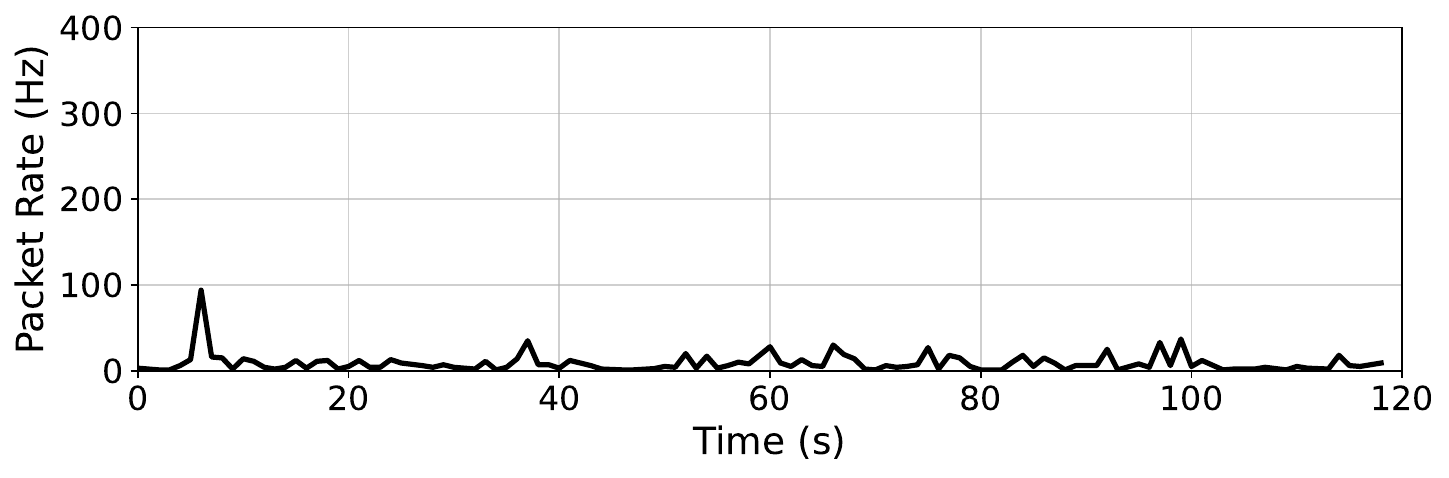}

  }

  \caption{Real-world traffic scenarios: (a) video streaming, (b) web browsing, (c) email send/receive, and (d) idle (no user activity).}
  \label{fig:realworldtraffic}
\end{figure}

To demonstrate variable traffic patterns in modern Wi-Fi networks, we measured 120-second packet-arrival traces under video streaming (YouTube), web browsing, email send/\allowbreak receive, and idle conditions. A packet-arrival recording lists the arrival time of every packet received in these 120 seconds. Fig.~\ref{fig:realworldtraffic}(a)--(d) shows the packet rates of these recordings, which are highly heterogeneous. Video streaming ranges from 1~Hz to 352~Hz, with an average of 67.10~Hz. Web browsing and email average 26.8~Hz and 22.8~Hz, respectively, and the idle condition averages 10~Hz.
\label{para:r1c2-trace-definition}

  The following subsections detail three key challenges for Wi-Fi sensing: low sampling rate, sampling rate generalisation, and fixed-length input.

\subsubsection{Low Sampling Rate}\label{subsec:lsr}

Fig.~\ref{fig:compare_SR_sig} illustrates the signal representation for the same motion (Push \& Pull) under various sampling rates and interval types. The black line represents the original 1000~Hz signal, while the red cross and blue circle markers represent downsampled signals with randomly and uniformly distributed sampling intervals, respectively.

Fig.~\ref{fig:compare_SR_sig}(a) shows that at 100~Hz, both uniform and random sampling still capture the signal's characteristic patterns reasonably well, with downsampled points closely following the original signal trajectory. However, as shown in Fig.~\ref{fig:compare_SR_sig}(b), reducing the sampling rate to 50~Hz causes severe information loss—the sparse sampling points fail to capture critical signal variations such as peaks and troughs, making it difficult to reconstruct the original motion pattern. This demonstrates that low sampling rates fundamentally limit the information available for motion recognition.

Recovering a motion-induced CSI component without temporal aliasing requires a sampling rate above twice its maximum Doppler frequency $f_{D,\max}$
~\cite{wang2015understanding}.
However, the objective of this work is motion recognition rather than recovery of the CSI sequence. The classifier therefore does not need every motion-induced frequency component. It only needs features that separate the target activity or gesture classes. Such features can include the coarse temporal evolution of the CSI, the duration of the motion, and the response pattern across subcarriers.

\subsubsection{Sampling interval} 
\begin{figure}[!t]
  \centering
  \subfloat[]{
    \includegraphics[width=1\linewidth]{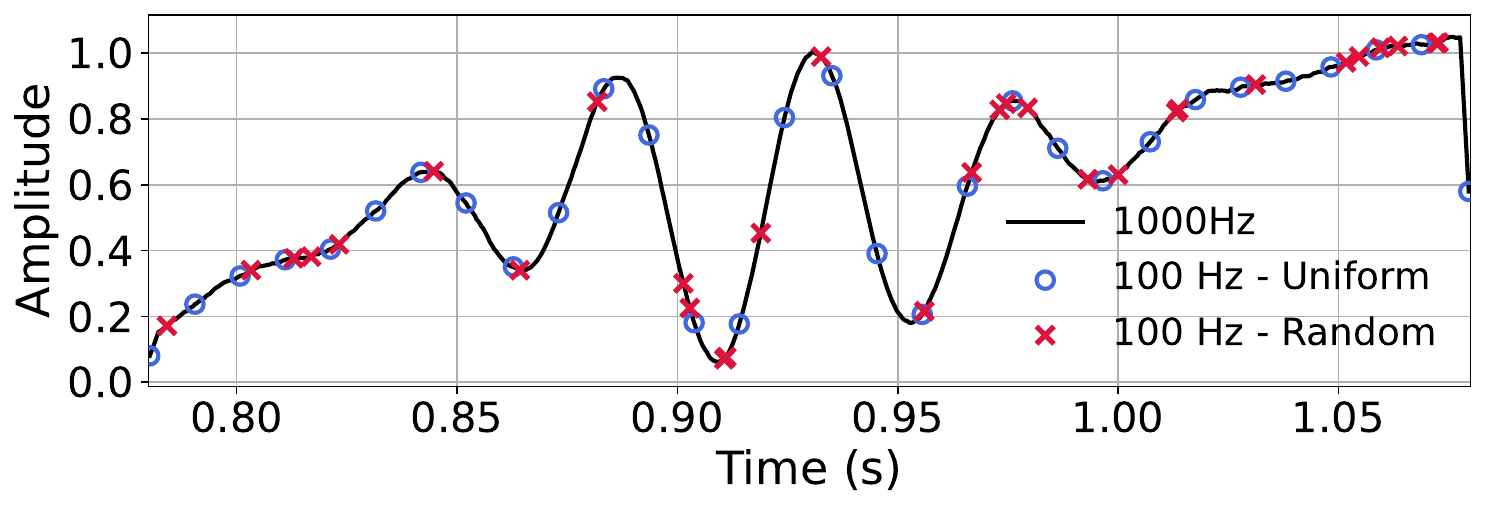}
    \label{fig:compare_SR_sig_high_rate}
  }

  \subfloat[]{
    \includegraphics[width=1\linewidth]{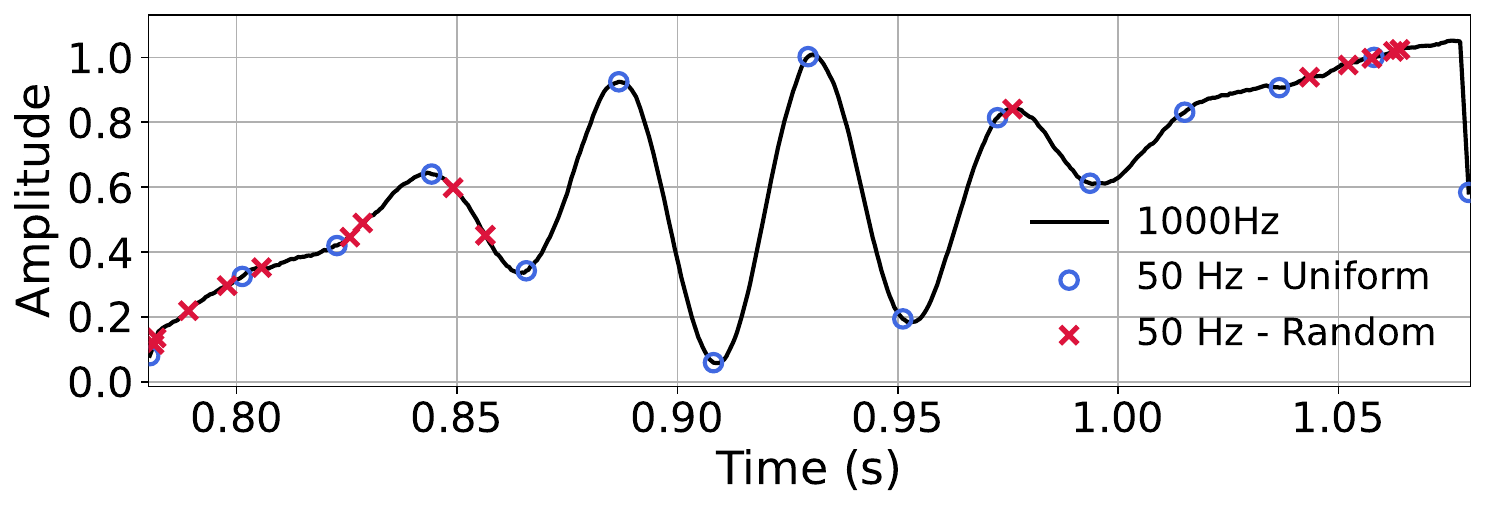}
    \label{fig:compare_SR_sig_low_rate}
  }

  \caption{The sensing signals for gestures of $Push \& Pull$ under different sampling rates and different types of sampling interval distributions. (a) 100 Hz. (b) 50~Hz. }
  \label{fig:compare_SR_sig}
\end{figure}

Sampling-interval variation is also often overlooked. Wi-Fi uses CSMA/CA with channel contention and random backoff, so packet arrivals are not uniformly spaced. As Fig.~\ref{fig:compare_SR_sig}(a) shows, at 100~Hz, randomly distributed intervals (red crosses) introduce variation compared with uniform intervals (blue circles), but the signal still preserves the original trend. At lower rates, however, random intervals compound the information loss: in Fig.~\ref{fig:compare_SR_sig}(b), the 50~Hz randomly sampled signal deviates more from the original than the uniformly sampled signal because sparse, randomly placed sampling points may miss critical features or oversample less informative regions.

To summarise, low sampling rates remove information and create feature variation, while random sampling intervals further distort the signal as the rate decreases. Wi-Fi sensing under low-rate traffic therefore faces a significant challenge.

\subsubsection{Sampling Rate Generalisation}\label{subsec:srg}
Existing Wi-Fi sensing systems often train deep models at a single sampling rate, so feature extraction is optimised for one temporal resolution and produces similar features $\mathbf{F}$ for the same motion only under that rate. These variations alter the signal's temporal resolution during inference, creating obstacles for models trained on a uniform sampling rate. As illustrated in Fig.~\ref{fig:compare_SR_sig}(a) and Fig.~\ref{fig:compare_SR_sig}(b), the representation of the same motion class can vary drastically across different sampling rates, compromising the model's ability to extract a consistent feature representation. Consequently, models trained on signals at a single sampling rate struggle to generalise to unseen sampling rates encountered during the inference phase. As the discrepancy between the observed signal and the training signals increases, the representation of the signal diverges further, leading to deteriorating recognition performance.

\subsubsection{Fixed-Length Input}\label{subsec:fix-len}
\begin{figure}[!t]
  \centering
  \includegraphics[width=1\linewidth]{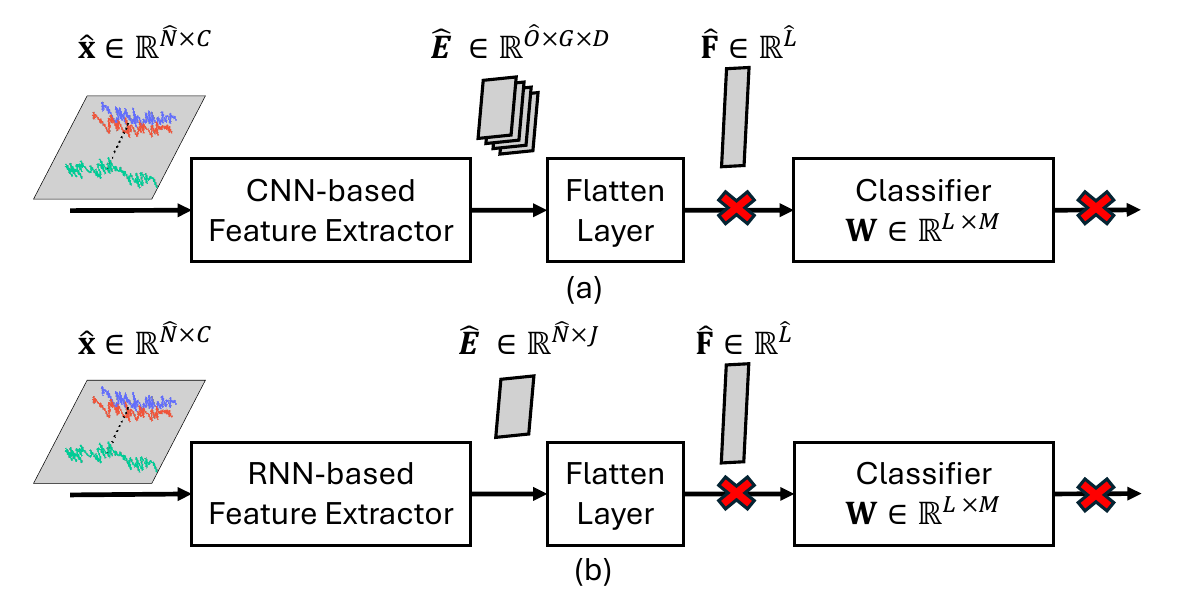}
  \caption{(a) Limitation of CNN with a flatten layer. (b) Limitation of RNN with a flatten layer.}
  \label{fig:limitations}
\end{figure}
Sampling-rate variation changes the input length, creating compatibility problems for models with flatten layers, such as those in Fig.~\ref{fig:model}(a) and Fig.~\ref{fig:model}(b). As shown in Fig.~\ref{fig:limitations}, when the number of sampling points changes from $N$ to $\hat{N}$, where $\hat{N} \neq N$, the input becomes $\hat{\mathbf{x}} \in \mathbb{R}^{\hat{N}\times C}$.
In the CNN-based model, the output matrix from the CNN-based feature extractor will become $\mathbf{\hat{E}} \in \mathbb{R}^{{\hat{O}}\times{G}\times{D}}$,  where $\hat{O} \neq O$. As a result, the flattened feature vector $\mathbf{\hat{F}}$ will have a different dimension, i.e., ${\hat{L}} = {\hat{O}}\times{G}\times{D}\neq {{L}}$. Therefore, the flattened feature is incompatible with the classifier's weights matrix $\mathbf{W} \in \mathbb{R}^{L \times M}$.
Similarly, in the case of the RNN-based model with a flatten layer, the output matrix from the RNN-based feature extractor will become $\mathbf{\hat{E}} \in \mathbb{R}^{\hat{N}\times J }$, which results in the flattened feature vector $\mathbf{\hat{F}}$ having a different dimension, i.e., ${\hat{L}} = \hat{N}\times J \neq {{L}} $, which will also result in the incompatibility problem.

The RNN model using the last hidden state for classification, i.e., shown in Fig.~\ref{fig:model}(c), can handle the variable input length signal, as the dimension of the feature vector $\mathbf{F}$ depends on the number of hidden units of the RNN-based feature extractor. The number of hidden units will not be affected by the input length.
However, to the best of the authors' knowledge, this capability has never been exploited in the existing Wi-Fi sensing works under variable input length scenarios. Although the work in~\cite{yang2023sensefi} used this method for classification, the input length was fixed. Therefore, the performance of the RNN under variable-length input in the Wi-Fi sensing context remains unexplored.

\section{System Overview}\label{sec:overview}
\begin{figure}[!t]
  \centering
  \includegraphics[width=3.4in]{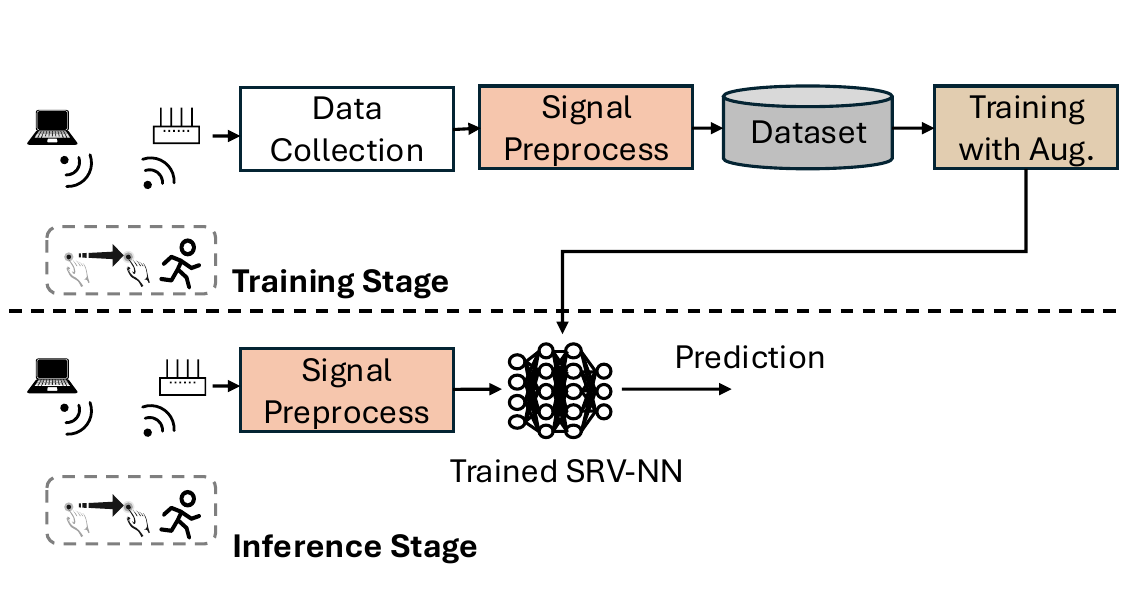}
  \centering
  \caption{System overview.}
  \label{fig:train_pip}
\end{figure}

Fig.~\ref{fig:train_pip} summarises the proposed system, which has training and inference stages. Its two main components are SRV-NN and dynamic sampling rate augmentation. SRV-NN is a transformer-based model that addresses fixed-length input limitations. Dynamic sampling rate augmentation broadens training diversity to improve sampling rate generalisation and low sampling rate accuracy.

In the training stage, CSI data captured during motion is preprocessed to correct hardware errors (Section~\ref{sec:signalprocess}) and used to train SRV-NN with dynamic sampling rate augmentation (Sections~\ref{sec:SRV-NN} and~\ref{sec:aug}). During inference, newly captured CSI undergoes the same preprocessing and is then classified by the trained SRV-NN.

\section{Signal Preprocessing}\label{sec:signalprocess}
CSI is susceptible to hardware anomalies, which typically manifest as anomalously large values and can skew subsequent analysis. We therefore correct these erroneous readings by interpolation before any further processing.

Given the CSI data $\mathbf{x} \in \mathbb{R}^{N \times C}$, values exceeding a predetermined threshold are treated as artefacts and zeroed to mark them invalid. Two passes then traverse the time and subcarrier dimensions in turn. The time-dimension pass interpolates each marked position from the temporally adjacent valid readings, unless the valid proportion of that time slice falls below 80\%, in which case the position is left as ``\textit{NaN}'' for the second pass. The subcarrier-dimension pass then interpolates the remaining positions along the subcarrier axis; if fewer than 80\% of the subcarriers at a timestamp are valid, that timestamp is deemed too corrupted to recover and is excluded from the sample. This procedure removes anomalous data and thereby improves CSI quality.

\section{Sampling Rate Versatile Neural Network Design}\label{sec:SRV-NN}

\begin{figure}[!t]
  \centering
  \includegraphics[width=3.4in]{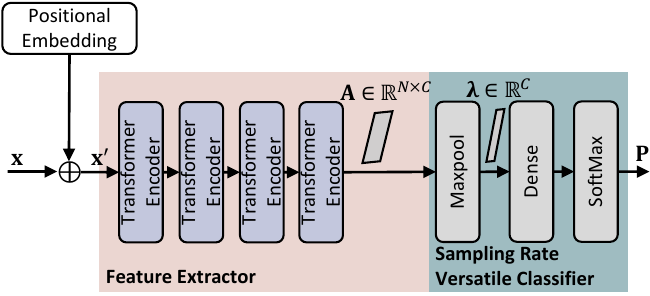}
  \centering
  \caption{Structure of the proposed learning model.}
  \label{fig:vc}
\end{figure}

To address the variable input length caused by sampling-rate variation (Section~\ref{subsec:fix-len}), we propose SRV-NN, which extracts length-invariant features for sensing across sampling rates. As shown in Fig.~\ref{fig:vc}, SRV-NN combines a transformer-based feature extractor for temporal dependency modelling with a sampling rate versatile classifier; positional encoding provides the temporal order information required by the transformer.

\subsection{Feature Extractor}\label{subsec:feature_extractor}
The feature extractor is designed to capture the temporal dependencies of the Wi-Fi sensing signal and extract meaningful representations that capture the unique characteristics and patterns associated with different motion classes.
We construct a feature extractor based on the encoder part of the transformer proposed in~\cite{vaswani2017attention}. We stack multiple encoders to form a
feature extractor, as shown in Fig.~\ref{fig:vc}.

\begin{figure}[!t]
  \centering
  \includegraphics[width=1.6in]{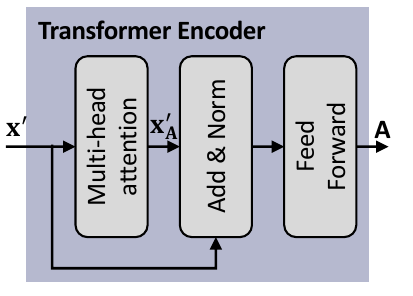}
  \caption{The structure of the transformer encoder}
  \label{fig:tel}
\end{figure}
The structure of the encoder is portrayed in Fig.~\ref{fig:tel}, which consists of a multi-head self-attention mechanism and a position-wise feed-forward network (FFN).
The core module of the transformer encoder is the multi-head self-attention, which takes the positionally encoded input $\mathbf{x}^\prime$ as input and aims to generate a hidden representation of inputs. By integrating self-attention to the transformer, the model can attend to different parts of the input sequence independently and extract various types of information.
The self-attention mechanism can be mathematically defined as
\begin{equation}\label{eq:mha}
  \operatorname{Attention}(\mathbf{Q},\mathbf{K},\mathbf{V})=\operatorname{softmax}\left(\frac{\mathbf{Q} \mathbf{K}^{\top}}{\sqrt{d}}\right) \mathbf{V},
\end{equation}
where $\mathbf{Q},\mathbf{K},\mathbf{V} \in \mathbb{R}^{N \times C}$ represent queries, keys and values, respectively, and the dot products are divided by $\sqrt{d}$, with $d=C$.
\label{para:r2c2-attention-scaling}
They are produced by linear transformation of the input $\mathbf{x'} \in \mathbb{R}^{N \times C}$, mathematically given as
\begin{equation}
  \mathbf{Q}=\mathbf{x}^\prime \mathbf{W_Q}, \mathbf{K}=\mathbf{x}^\prime \mathbf{W_K}, \mathbf{V}=\mathbf{x}^\prime \mathbf{W_V},
\end{equation}
where $\mathbf{W_Q},\mathbf{W_K},\mathbf{W_V} \in \mathbb{R}^{C \times C }$ are the linear layers. Based on the self-attention mechanics, multi-head attention is an extension of the attention mechanism that introduces multiple sets of attention computations independently using (\ref{eq:mha}). %
Each attention head of multi-head attention is associated with learned parameters and can attend to different patterns or features. These attention heads capture different aspects of the input, allowing the model to learn more complex and nuanced representations. The multi-head attention can be represented mathematically as
\begin{equation}\label{eq:multihead}
  \operatorname{MultiHead Attention}(\mathbf{Q}, \mathbf{K}, \mathbf{V})=\left[ { \beta }_1 ; \ldots ;  { \beta }_Z\right] \mathbf{W_U},
\end{equation}
where $Z$ is the number of attention heads and $\beta_{i}=\operatorname{Attention}\bigl(\mathbf{x}^\prime \mathbf{W_Q}^i,\allowbreak \mathbf{x}^\prime \mathbf{W_K}^i,\allowbreak \mathbf{x}^\prime \mathbf{W_V}^i\bigr)$ is the output of the $i$-th head. The matrices $\mathbf{W_Q}^i,\mathbf{W_K}^i,\mathbf{W_V}^i \in \mathbb{R}^{C \times C/Z}$ are the $i$-th blocks of $C/Z$ columns of $\mathbf{W_Q}$, $\mathbf{W_K}$ and $\mathbf{W_V}$. The $C$ features are thus split evenly among the $Z$ heads, so $Z$ must divide $C$. The outputs of the heads are concatenated into an $N \times C$ matrix and multiplied by $\mathbf{W_U} \in \mathbb{R}^{C \times C}$ to obtain the output of the multi-head self-attention module, denoted as $\mathbf{x_{A}^{\prime}}$.
\label{para:r2c2-mha-heads}
A residual connection~\cite{he2016deep} adds the attention output to the input $\mathbf{x}^\prime$, and layer normalisation is applied to this sum.
The final output of an encoder (attention score), $\mathbf{A}$, will be computed by a two-layer feedforward network:
\begin{equation}
  \mathbf{A} = \text{FFN}(\text{Norm}(\mathbf{x_{A}^{\prime}}+\mathbf{x}^\prime)),
\end{equation}
where Norm($\cdot$) is the layer normalisation.

In practice, the first self-attention layer of the transformer takes the sequential time-series data $\mathbf{x}^\prime \in \mathbb{R}^{N\times C}$ as input. The output of the feature extractor, also known as attention score, $\mathbf{A} \in \mathbb{R}^{N \times C}$, matches the size of the input sequence.

\subsection{Sampling Rate Versatile Classifier}
A sampling rate versatile classifier is designed to manage the inherent variation in the sampling rates of input signals, facilitating motion classification irrespective of the input sequence length. As discussed in the previous section, the first dimension $N$ of the output $\mathbf{A}$ of the feature extractor is variable, depending on the signal length which may vary dramatically in practice. To handle such a variable length tensor, a max pooling layer pools over the time dimension, i.e., the $N$ dimension. It produces a vector $\mathbf{\lambda} \in \mathbb{R}^{C}$, whose dimension does not change as the input sequence length $N$ changes.

Following the max pooling operation, a fully connected layer with a weight matrix $\mathbf{W_C} \in \mathbb{R}^{C \times M}$ maps $\mathbf{\lambda}$ to an $M$-dimensional space, one dimension per class. A softmax function is then applied to the outputs to yield a probability distribution $\mathbf{P}$ over the $M$ classes.
\begin{equation}
  \mathbf{P} = \text{softmax}(\mathbf{\lambda}^{\top} \mathbf{W_C} + \mathbf{B_C}),
\end{equation}
where $\mathbf{W_C}$ is the weight matrix of the sampling rate versatile classifier and the $\mathbf{B_C}$ is the corresponding bias.

The design of the classifier is critical for sampling rate generalisation. Transformer positional encoding associates feature representations with sequence indices rather than physical time, so a change in the sampling rate both moves the same temporal event to a substantially different position and alters the number of feature vectors that carry it. Representations that rely on absolute position, or on the sequence length, therefore generalise poorly across variable-length CSI sequences. To mitigate this issue, the sampling rate versatile classifier applies max pooling over the temporal dimension to preserve the most salient motion patterns while suppressing both dependences, thereby enabling consistent classification across the wide range of input lengths encountered in practical Wi-Fi sensing systems. Max pooling is deliberately adopted because the maximum of each feature is unchanged when the feature vector that holds it moves to another position, or when feature vectors with smaller values are added or removed. Average pooling, the most common standard temporal pooling, instead divides by the number of feature vectors, so its output changes whenever the sampling rate adds or removes feature vectors.
\label{para:r3c3-srv-classifier}

\section{Dynamic Sampling Rate Augmentation}\label{sec:aug}

\begin{figure}[!t]
  \centering
  \includegraphics[width=0.82\linewidth]{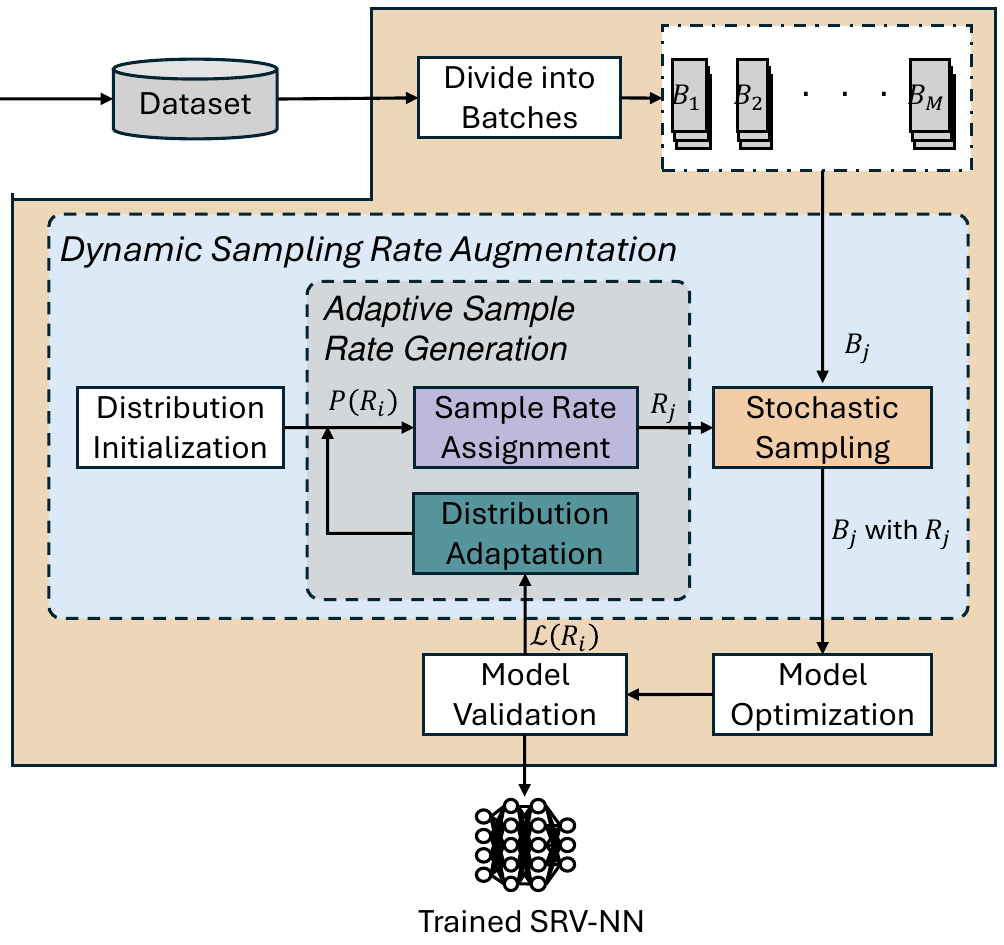}
  \caption{The overall workflow of training with augmentation.}
  \label{fig:aug}
\end{figure}

To address the challenges in Sections~\ref{subsec:lsr} and~\ref{subsec:srg}, we propose dynamic sampling rate augmentation, shown in Fig.~\ref{fig:aug}. In each training cycle, the dataset is divided into batches, each batch is augmented before model training, and validation across sampling rates is used to adjust the rate distribution for the next cycle.
This augmentation method is comprised of two key components:
\begin{itemize}
  \item \textbf{Adaptive sampling rate generation} augments the diversity of the sensing signal with respect to its sampling rate.

  \item \textbf{Stochastic sampling} addresses the challenges associated with low sampling rates and the randomness of sampling points.
\end{itemize}

The proposed augmentation differs from those in Section~\ref{subsec:general_timeseries} in two ways. First, for each training batch, it draws a sampling rate and keeps a random subset of the CSI sampling points at that rate. The training sequences therefore differ in length and have irregular sampling intervals. Second, the distribution of the sampling rates is adapted during training according to the validation loss (Section~\ref{sec:ushape}).
\label{para:r1c1-aug-difference}

\subsection{Adaptive Sampling Rate Generation}
Adaptive sampling rate generation has two elements: sampling rate assignment and distribution adaptation. It assigns different training rates to different batches and adaptively concentrates on rates that are harder to generalise, thereby improving learning efficiency across a wide rate range.

During training, this augmentation enriches sampling-rate and sampling-point diversity by resampling the original signal from rate $R$ to $R_j$, where $R>R_j$, using the stochastic sampling in Section~\ref{sec:ssa}. The initial distribution is uniform over the bounded range $[R_{low}, R_{upper}]$, where each rate has probability $P(R_j)$ and is used by the sampling rate assignment block in the first epoch.

\subsubsection{Sampling Rate Assignment}
In each training epoch, the probability distribution will be fed into the adaptive sampling rate generation module, and the sampling rate assignment block will draw a sampling rate $R_j$ for each batch $B_j$ based on the probability distribution,
\begin{equation}
  R_j \sim \mathcal{U}(R_{\text{low}}, R_{\text{upper}}),
\end{equation}
where $\sim$ denotes that $R_j$ is drawn from a probability distribution $\mathcal{U}$.
Varying the sampling rate during training simulates the rate variation encountered in practice, which prevents the model from overfitting to a single rate and encourages it to learn representations that transfer to rates unseen during training.

\subsubsection{Distribution Adaptation}\label{sec:ushape}
Drawing $R_j$ from a uniform distribution overlooks the distinct difficulty of each rate: high rates carry richer information and are easily learned, whereas low rates offer less information and larger representation variation. Balanced performance across the rate spectrum therefore requires the training focus to follow this difficulty rather than treating all rates equally. To this end, we propose a distribution adaptation method that adjusts the probability distribution over the course of training, prioritising the rates at which the model currently performs worst.

The core of the probability distribution adaptation method is to adjust the selection probability of each rate $R_i$ based on the model's performance at that sampling rate. The adjustment is governed by a key equation that recalculates the probability $P(R_i)$ as follows
\begin{equation}\label{sqn:adj}
  P(R_i) \leftarrow P(R_i) + \Delta P(R_i),
\end{equation}
where $\Delta P(R_i)$ is the adaptation made to the existing probability $P(R_i)$. The probability distribution $\mathcal{U}(R_{\text{low}}, R_{\text{upper}})$ is initialised as a uniform distribution. The adaptation is implemented for each epoch, and the $\Delta P(R_i)$ is calculated based on $\mathcal{L}(R_i)$, which represents the aggregation of losses corresponding to each sampling rate. This collection of losses is acquired during the validation phase, wherein the model's loss for each sampling rate is  recorded. Subsequently, the $\Delta P(R_i)$ is determined as follows:
\begin{equation}\label{sqn:norm}
  \Delta P(R_i) = P(R_i) \cdot \left( \frac{\mathcal{L}(R_i) - \mathcal{L}_{\min}}{\mathcal{L}_{\max} - \mathcal{L}_{\min}} \right) \cdot \alpha,
\end{equation}
where $\mathcal{L}_{\min}$ and $\mathcal{L}_{\max}$ are the minimum and maximum losses observed across all rates, respectively. The $\alpha$ parameter represent a distribution adaptation learning rate that controls the magnitude of the adjustment, allowing for tuning based on the specific characteristics of the training data and the model. After applying the adjustments in (\ref{sqn:adj}), the probabilities $P(R_i)$ will be normalised, given as
\begin{equation}\label{eqn:prob_norm}
  P(R_i) \leftarrow \frac{P(R_i)}{\sum_{j} P(R_j)}.
\end{equation}

Distribution adaptation therefore focuses training on sampling rates that are harder to generalise.

\subsection{Stochastic Sampling}\label{sec:ssa}
On each training batch with an assigned batch sampling rate $R_b$, we downsample the original sensing signal from $R$ into $R_b$ by selecting $N_b$ sampling points. The $N_b$ is computed as:

\begin{equation}
  N_b = \frac{NR_{b}}{R}.
\end{equation}

During the sampling point selection process, instead of selecting sampling points with equal sampling intervals ($\Delta t$), we selected sampling points with random sampling intervals. Because the time duration of the sample $\mathbf{x}$ is $T$, we have
\begin{equation}
  T = \sum^{N_b}_{i=1} \Delta t_i,
\end{equation}
where $\Delta t_i$ is the sampling interval between the $i$-th and $(i+1)$-th sampling points.

This approach improves SRV-NN robustness to randomly distributed sampling points. By randomising the selected points for each training sample, it exposes the model to diverse sampling-interval patterns, mitigates overfitting to fixed intervals, and improves generalisation under low and irregular sampling rates.
One might consider upsampling lower-rate data to a fixed high rate instead of downsampling. However, low-rate signals inherently contain less information, and upsampling cannot recreate what was lost during low-rate sampling, so training on such artificially upsampled data would provide incomplete representations, which is particularly problematic given the variable sampling rates and intervals of real-world scenarios. Downsampling instead leverages information-complete high-rate data and projects it to various lower rates with diverse sampling patterns, enabling the model to learn robust rate-invariant features from information-rich representations.

\begin{table*}[!t]
  \centering
  \caption{Summary of datasets}
  \begin{tabular}{cccccccccc}
    \toprule
    & \textbf{\# samples} & \textbf{\# Classes} & \textbf{Duration} & \textbf{SR} & \textbf{BW} & \textbf{Channel} & \textbf{\# Ant} & \textbf{\# Subs/Ant} & \textbf{Device} \\
    \midrule
    \textbf{SRV Activity} & 1500   & 5     & 2s    & 600 Hz & 80 MHz & 122   & 4     & 242   & AC86 \\
    \textbf{SRV Gesture} & 1200  & 7     & 1s    & 600 Hz & 80 MHz & 58    & 4     & 242   & AC86 \\
    \textbf{SHARP} & 6300  & 9     & 2s    & 173 Hz & 80 MHz & 42    & 4     & 242   & AC86 \\
    \textbf{Widar} & 3000  & 6     & 2s    & 1000 Hz & 20 MHz & 165   & 3     & 30    & Intel5300 \\
    \bottomrule
  \end{tabular}%
  \label{tab:dataset}%
\end{table*}%

\section{Experimental Evaluation}\label{sec:experiments}

\subsection{Experiment Set-up}

\begin{figure}
  \centering
  \includegraphics[width=1\linewidth]{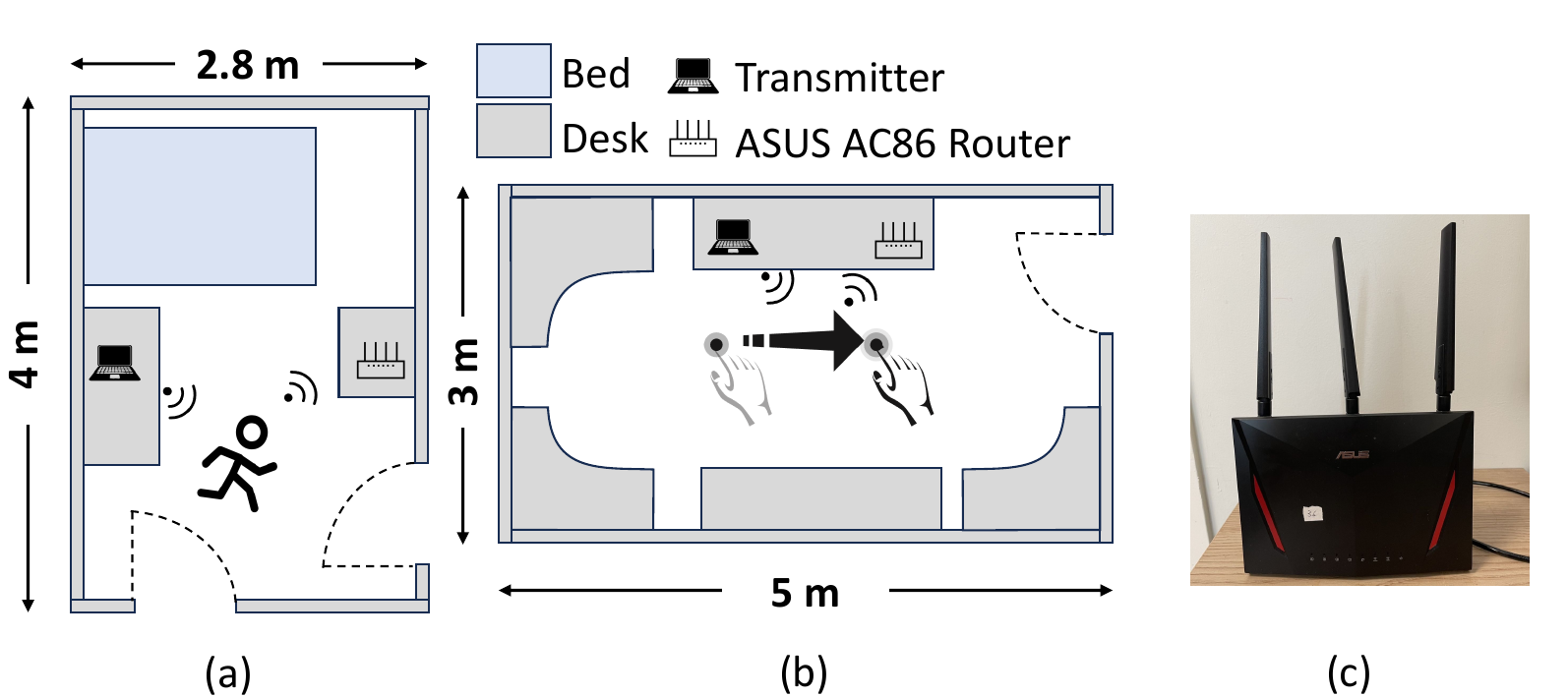}
  \caption{Floor plan and device of data collection. (a) The floor plan of SRV Activity dataset collection. (b) The floor plan of SRV Gesture dataset collection. (c) Receiver (ASUS AC86 router)}
  \label{fig:floorplan}
\end{figure}

\subsubsection{Dataset Description}\label{sec:dataset_desc}

  We collected two sensing datasets, denoted as SRV Activity and SRV Gesture. The experimental setup consisted of one transmitter (MacBook) and one Wi-Fi receiver (ASUS AC86 router equipped with four antennas: three external and one internal antenna). Probing packets were transmitted using iperf, a network traffic generation and performance measurement tool, at a rate of 600 Hz, and the Nexmon CSI tool~\cite{nexmon} was employed to extract CSI measurements from the received packets. IEEE 802.11ac with 80 MHz bandwidth was used, with 242 subcarriers available per antenna. The floor plan and router used for the experiments are shown in
Fig.~\ref{fig:floorplan}.

  The SRV Activity dataset was collected in a typical bedroom environment (4m × 2.8m) furnished with desks, chairs, computers, and bed. Data collection was conducted on Channel 122 with one individual. For each sample, CSI was recorded for 2~seconds while the participant performed one target activity. The five measured classes were \textit{Walking}, \textit{Jumping}, \textit{Waving Hand}, \textit{Picking Up}, and an \textit{Empty} room baseline (no human motion). This resulted in a total of 1,500 samples across 5 classes, with 300 samples per activity.

  The SRV Gesture dataset was collected in an office environment (3m × 5m) containing typical office furniture including desks, chairs, desktop, monitors, and metal cabinets. The experimental configuration followed the same spatial setup as the activity dataset, with data collection conducted on Channel 58, and the individual as the activity dataset, 6 hand gestures were performed plus an empty room condition (\textit{Up \& Down}, \textit{Pull \& Push}, \textit{Drawing O}, \textit{Drawing Infinity}, \textit{Drawing N}, \textit{Drawing W}, \textit{Empty}). Each gesture sample lasted 1 second, resulting in 1,200 samples across 7 classes.

  Both the transmitter and receiver were positioned at a height of 1.1m. The participants were instructed to perform each activity or gesture naturally within the designated sensing area, with sufficient rest intervals between samples to ensure data quality.
  Detailed specifications of both datasets are provided in
Table~\ref{tab:dataset}.
All data collection procedures were approved by the departmental ethical committee of the Department of Electrical Engineering and Electronics, University of Liverpool, UK, and informed consent was obtained.

We also utilised two publicly available datasets for benchmarking, i.e., SHARP
~\cite{meneghello2022sharp}
and Widar
~\cite{widar} datasets.
\begin{itemize}
  \item SHARP activity dataset was collected in two different environments, i.e., a bedroom and a living room, with a sampling rate of 173 Hz and three individuals. It encompasses eight activities ($Walking$, $Running$, $Jumping$, $Sitting$, $Standing$, $Sitting-down$, $Standing-up$, and $Doing-arm-gym$) and an empty room without motion, The data collection was conducted via Channel 42 with an 80 MHz bandwidth. An ASUS AC86 router with Nexmon CSI tool was used.

  \item Widar gesture dataset was collected in a classroom with a sampling rate of 1000 Hz and one individual. The dataset consists of six gestures: $Push \& Pull$, $Sweep$, $Clap$, $Slide$, $Draw-Zigzag$, and $Draw-N$. We used a dataset containing 3,000 samples.  The data collection was performed using Channel 165 with a 20~MHz bandwidth. An Intel 5300 Network Interface Card
~\cite{Halperin_csitool} was used.
\end{itemize}

The samples in all four datasets were collected in multiple independent sessions at different times. Training and test samples were acquired in separate sessions.
\label{para:r1c4-sessions}
A summary of all the datasets used in this paper is given in
Table.~\ref{tab:dataset}.

\subsubsection{Device Information and CSI Extraction Tools}
In this study, the datasets were collected using two different devices: the ASUS AC86 Router and the Intel NIC 5300.

The SRV Activity, SRV Gesture, and SHARP datasets were acquired using an ASUS AC86 router, equipped with the Nexmon CSI tool~\cite{nexmon}. During the data collection, the router was configured to operate using an 80~MHz bandwidth and the IEEE 802.11ac protocol, leveraging 4 antennas with 242 valid subcarriers for each antenna.

The Widar dataset was obtained using an Intel  5300 NIC, which was set up with the Linux IEEE 802.11 CSI tool~\cite{Halperin_csitool} to record CSI effectively. For precise data logging, the devices were configured in ``monitor'' mode, operating using channel 165 with 20 MHz bandwidth, and 3 antennas with 30 subcarrier groups for each antenna.

\subsubsection{Model Training and Testing}\label{sec:cfg}
The neural network was implemented using PyTorch, a powerful and widely-used deep learning framework. The training process was conducted on a workstation equipped with an Nvidia GeForce RTX 3060 GPU and 64~GB of RAM.

  The same training method and hyperparameters were used for all datasets, except for the number of attention heads. We used a batch size of 16, the Adam optimiser and an initial learning rate of $1\times10^{-5}$. The learning rate was reduced by a factor of 0.1 after 10 epochs without improvement in the validation loss, and training stopped after 20 epochs without improvement. The feature extractor stacks six encoders, as in the transformer architecture reported in~\cite{vaswani2017attention}. As explained in Section~\ref{subsec:feature_extractor}, the number of heads $Z$ in (\ref{eq:multihead}) must divide the feature dimension $C$. We used $Z=11$ heads ($C/Z=88$) for SRV Activity, SRV Gesture and SHARP, where $C=968$, and $Z=3$ heads ($C/Z=30$) for Widar, where $C=90$.

The train/test split is made on the original samples at their native sampling rates, before any downsampling. Dynamic sampling rate augmentation is applied only to training batches, so no downsampled version of a test sample is included in a training batch. The test set at each sampling rate in Table~\ref{tab:testsr} is built by downsampling every test sample to that rate with randomly selected sampling points, so the test samples have irregular sampling intervals.
\label{para:r1c4-protocol}

  \subsubsection{Evaluation Metrics}\label{sec:eval_metrics}
  In this work, we propose a dual-metric evaluation framework specifically designed for variable sampling rate scenarios: (1) \textit{average accuracy} across diverse sampling rates to ensure overall performance, and (2) \textit{variance} (or standard deviation) across sampling rates to measure model robustness and consistency. A model that achieves high accuracy at one sampling rate but fails dramatically at others is impractical for deployment in real-world Wi-Fi networks. This evaluation paradigm better reflects the practical requirements where sensing systems must operate reliably under unpredictable network conditions without requiring dedicated high-rate data streams.
  Formally, let $\mathcal{R} = \{R_1, R_2, ..., R_n\}$ denote the set of $n$ sampling rates evaluated for a given dataset, and let $A(R_i)$ represent the classification accuracy achieved by a model at sampling rate $R_i$. The average accuracy measures the overall performance across all sampling rates:
  \begin{equation}
    \bar{A} = \frac{1}{n}\sum_{i=1}^{n} A(R_i)
  \end{equation}
  The variance quantifies the stability and robustness of the model across different sampling rates:
  \begin{equation}
    \text{Var}(A) = \frac{1}{n}\sum_{i=1}^{n} \left(A(R_i) - \bar{A}\right)^2
  \end{equation}
  A model suitable for practical deployment in variable sampling rate scenarios should achieve both high average accuracy $\bar{A}$ and low variance $\text{Var}(A)$ (or low standard deviation $\sigma(A)$), indicating consistent performance regardless of network traffic conditions.

\subsection{Selection of Learning Models}\label{sec:model_selection}

\begin{table*}[!t]
  \centering
  \caption{Selected sampling rates for different datasets}
  \begin{tabular}{ccccccccccccc}

    \toprule
    SRV Activity & 5 Hz  & 10 Hz & 20 Hz & 30 Hz & 40 Hz & 50 Hz & 100 Hz & 200 Hz & 300 Hz & 400 Hz & 500 Hz & 600 Hz \\
    SRV Gesture & 5 Hz  & 10 Hz & 20 Hz & 30 Hz & 40 Hz & 50 Hz & 100 Hz & 200 Hz & 300 Hz & 400 Hz & 500 Hz & 600 Hz \\
    SHARP  & 5 Hz  & 10 Hz & 20 Hz & 40 Hz & 60 Hz & 80 Hz & 90 Hz & 100 Hz & 120 Hz & 140 Hz & 160 Hz & 173 Hz \\

    Widar  & 5 Hz  & 10 Hz & 20 Hz & 30 Hz & 40 Hz & 50 Hz & 100 Hz & 200 Hz & 400 Hz & 600 Hz & 800 Hz & 1000 Hz \\
    \bottomrule
  \end{tabular}%
  \label{tab:testsr}%
\end{table*}%

\begin{figure}[!t]
  \centering
  \includegraphics[width=1\linewidth]{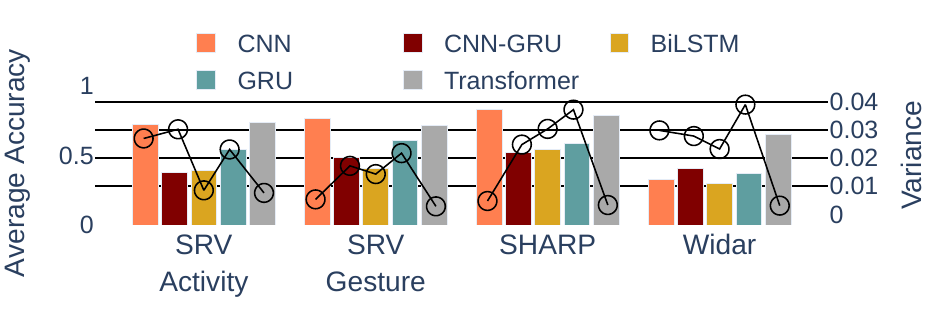}
  \caption{The performance of different models on different datasets. The bar plot is the average accuracy and the line plot is the variance.}
  \label{fig:model_performance}
\end{figure}

This section evaluates the proposed transformer against representative feature-extraction architectures under variable sampling rates, with the aim of selecting the backbone used in the subsequent experiments.

The comparison includes the GRU-based model from~\cite{yang2023sensefi}, the bidirectional LSTM-based model from~\cite{chen2018wifi}, the CNN-based model from~\cite{yin2022fewsense}, and the hybrid CNN-GRU model from~\cite{widar}. Together, these baselines cover the recurrent, convolutional, and hybrid paradigms widely used in Wi-Fi sensing, enabling a systematic assessment of backbone choice under sampling-rate variation.
In this comparison, dynamic sampling rate augmentation was disabled for all models. The CNN used the same sampling rate versatile classifier as SRV-NN. GRU, BiLSTM and CNN-GRU classified from their final hidden state.
\label{para:r1c5-controlled-backbone-comparison}
All models were trained at 100~Hz and evaluated across the dataset-specific sampling rates listed in Table~\ref{tab:testsr}. The convolutional layers of the CNN and CNN-GRU need a minimum input length, so their average accuracy and variance in Fig.~\ref{fig:model_performance} are computed without the lowest test sampling rates: 5 and 10~Hz for the CNN and 5~Hz for the CNN-GRU, plus 20 and 10~Hz, respectively, on SRV Gesture.

Fig.~\ref{fig:model_performance} reports average accuracy and accuracy variance across the sampling rates in Table~\ref{tab:testsr}; the former measures overall recognition performance, while the latter reflects stability across sampling rates.
With regard to average accuracy over all datasets, the proposed transformer feature extractor ranked best, with an average accuracy of 72\%, followed by CNN of 66\%. The GRU, CNN-GRU, BiLSTM based feature extractors reveal lower average accuracies of 53\%, 45\%, and 41\%, respectively. The transformer architecture also achieved lowest variance (0.0041) over all the tested datasets, followed by the BiLSTM family (0.013). Conversely, the CNN, CNN-GRU, and GRU exhibited larger variances of 0.0196, 0.02704, and 0.03, respectively. Although the CNN occasionally achieved slightly higher accuracy than the transformer in some cases, its substantially larger variance indicates much lower stability under sampling rate variation; therefore, according to the dual-metric evaluation in the previous section, the transformer remains the better model choice.

These results guide model selection under sampling-rate variation: both average accuracy and variance must be considered. Since the transformer achieves the highest average accuracy and lowest variance, the proposed transformer-based SRV-NN is used in the remaining experiments.

\subsection{Investigating Sampling Rate Effects on Wi-Fi Sensing with Conventional Training}\label{sec:srimpact}
To examine how the sampling rate impacts Wi-Fi sensing models trained with the conventional method, i.e., at a single fixed rate, we trained 12 SRV-NN models for each dataset, one per training rate specified in Table~\ref{tab:testsr}, and evaluated every model across all the test sampling rates listed in the same table. For example, the model trained on the SRV Activity dataset at 5~Hz is tested across rates ranging from 5~Hz to 600~Hz.

\begin{figure*}[!t]
  \centering
  \subfloat[]{
    \includegraphics[width=0.25\linewidth]{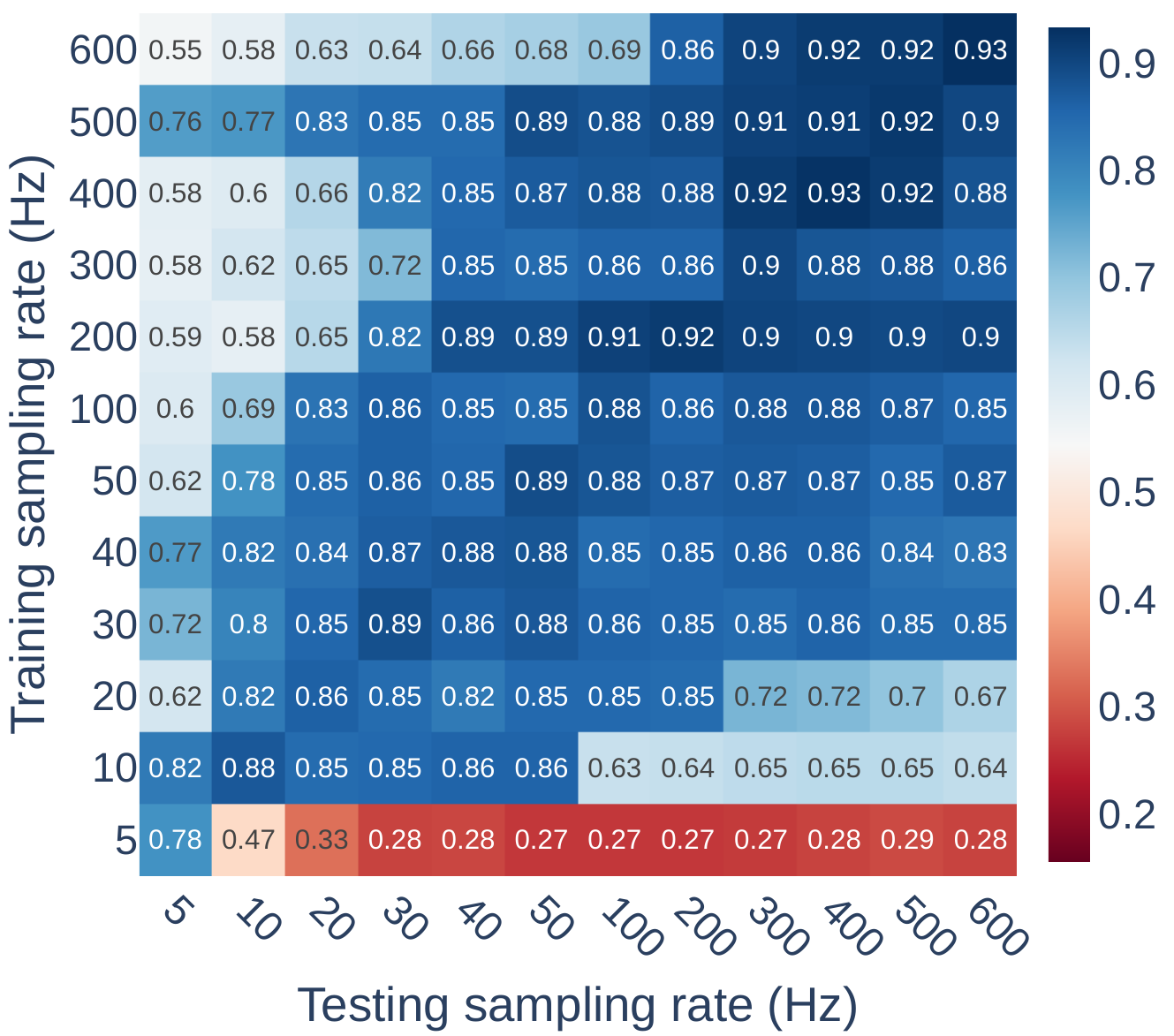}
  }
  \subfloat[]{
    \includegraphics[width=0.25\linewidth]{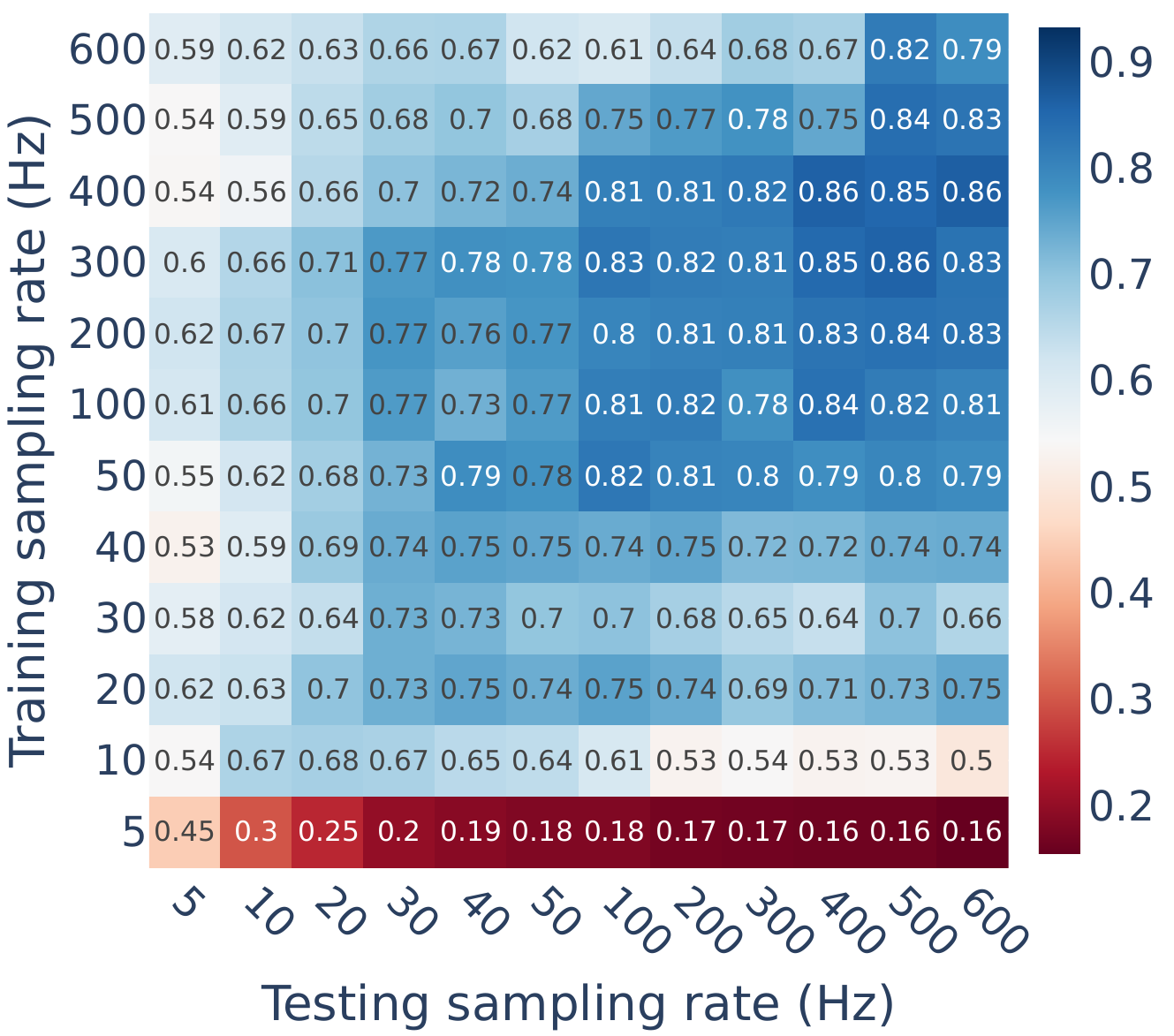}
  }
  \subfloat[]{
    \includegraphics[width=0.25\linewidth]{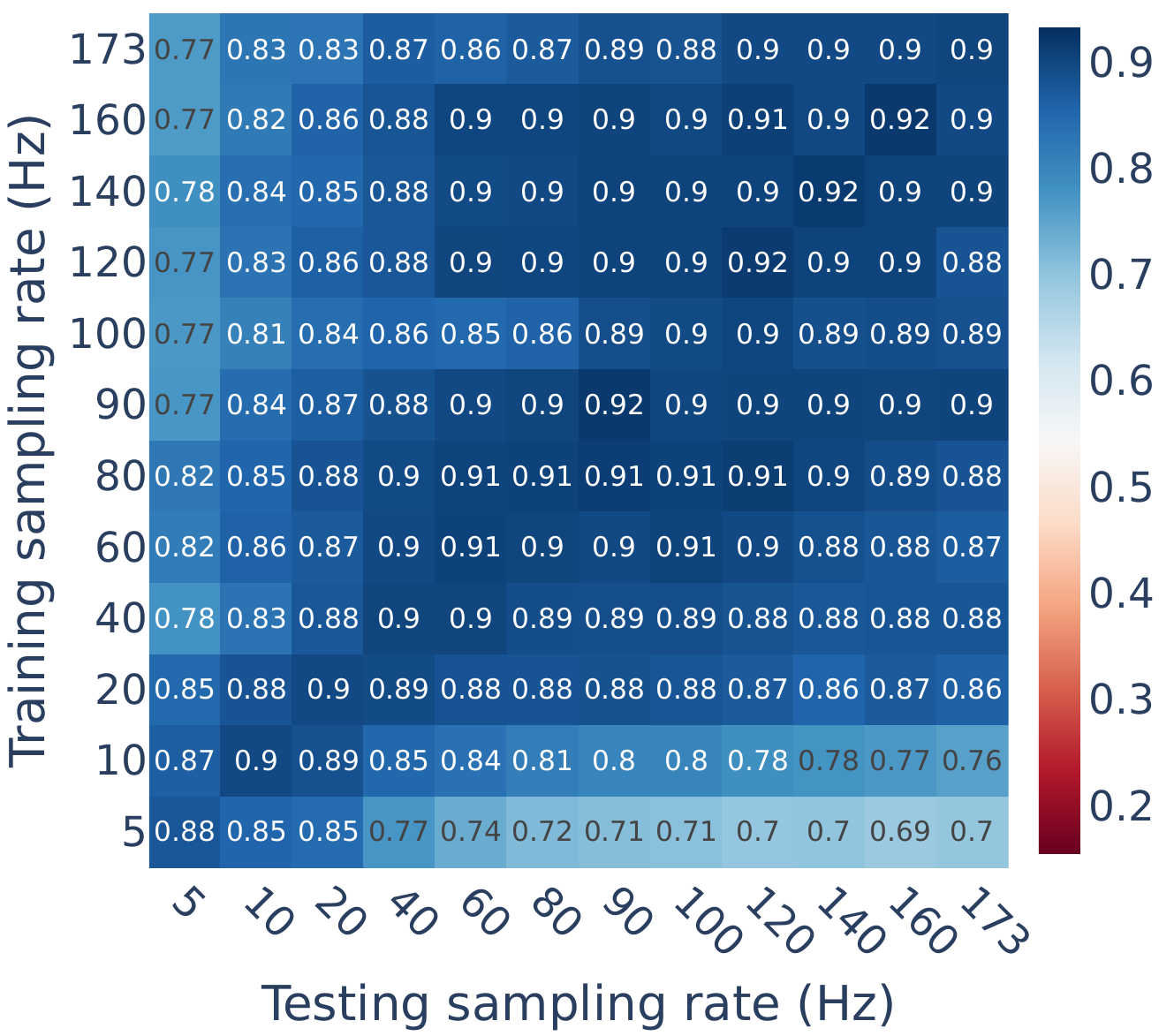}
  }
  \subfloat[]{
    \includegraphics[width=0.25\linewidth]{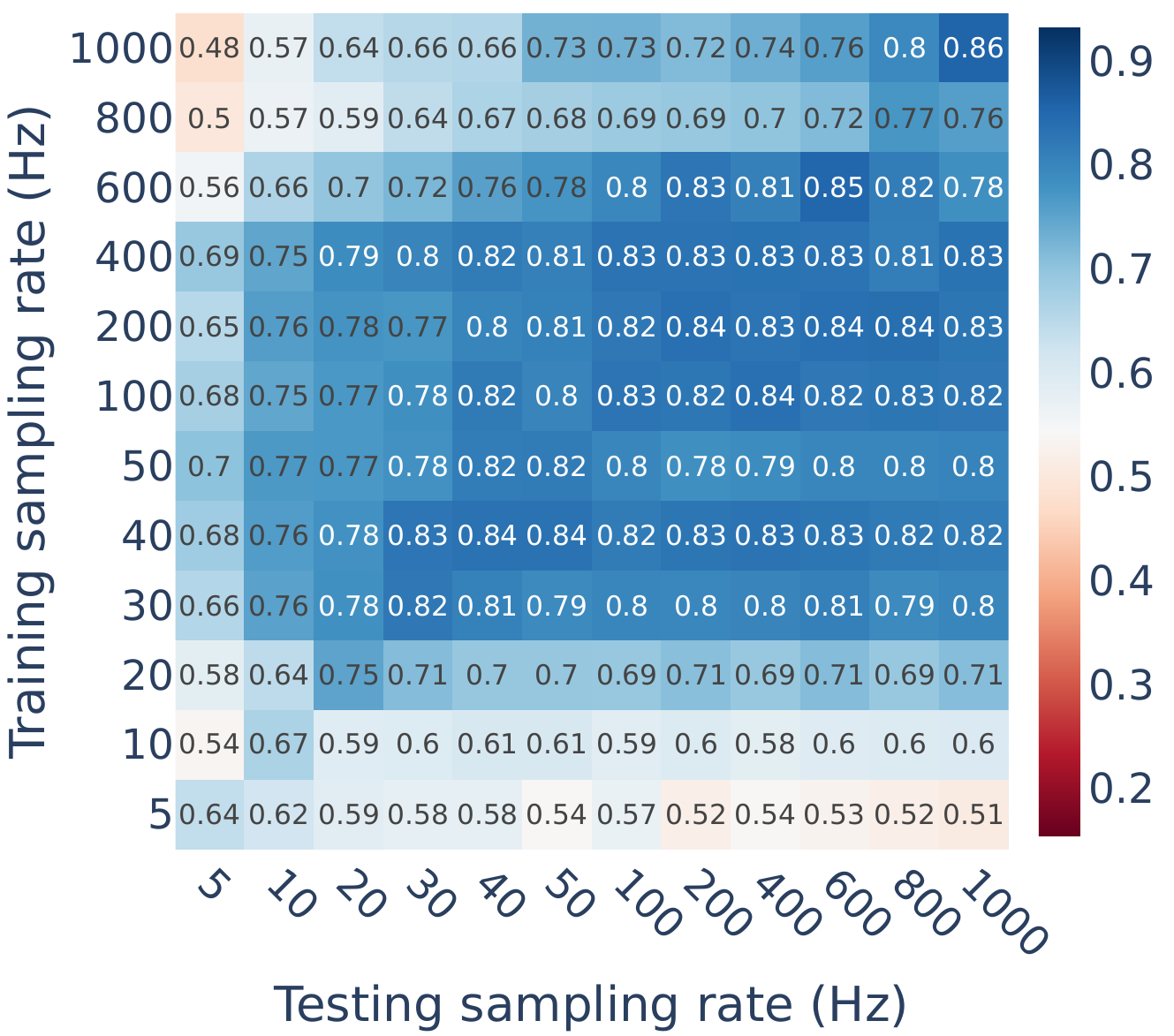}
  }
  \caption{Analysing the effectiveness of models trained and tested with diverse sampling rates on distinct datasets. (a) SRV Activity dataset. (b) SRV Gesture dataset. (c) SHARP dataset. (d) Widar dataset. }
  \label{fig:heatmap}
\end{figure*}

Fig.~\ref{fig:heatmap} summarises the relationship between training and testing rates using heatmaps, where the y-axis indicates the training sampling rate and the x-axis indicates the test sampling rate.

The results show a notable correlation between their training and testing rates. The highest accuracy is consistently observed when models are both trained and tested at similar rates, as clearly illustrated by the enhanced accuracy along the heatmap's diagonal. For SRV Activity, SRV Gesture, SHARP, and Widar, the average accuracies along the diagonal were 88.8\%, 75.0\%, 90.5\%, and 80.0\%, respectively. These figures are significantly superior to the average off-diagonal accuracies, which stand at 76.1\%, 66.5\%, 85.8\% and 73.4\% for each dataset accordingly.

Furthermore, as the discrepancy between the training and testing sampling rates increases, there is a marked decline in the accuracy of the corresponding models. For example, on SRV Activity, the model trained at 5~Hz drops from 78.1\% to 28.0\% when tested at 600~Hz. Conversely, the model trained at 600~Hz drops from 93.6\% to 55.1\% when tested at 5~Hz. The same trends were also shown in other datasets.

These findings highlight the two critical effects of the sampling rate on Wi-Fi sensing systems:
\begin{itemize}
  \item The performance of the Wi-Fi sensing model demonstrated low accuracy in low-sampling scenarios.
  \item The model that has been trained using a constant sampling rate tends to overfit to the specific rate used during training. The optimal model performance is achieved when the training and testing sampling rates are closely matched; deviations from the training rate result in a marked decline in accuracy.
\end{itemize}
Given the real-world constraints discussed in Section~\ref{sec:motivation}, it is necessary for models that excel in sampling rate generalisation to maintain consistent performance across a broad spectrum of rates.

\subsection{Evaluation of Proposed Sensing System}
This section evaluates SRV-NN trained with dynamic sampling rate augmentation and ablates the roles of stochastic sampling and distribution adaptation in sampling rate generalisation.

\subsubsection{Dynamic Sampling Rate Augmentation}\label{sec:aug_results}

\begin{figure*}[!t]
  \centering
  \subfloat[]{
    \includegraphics[width=0.48\linewidth]{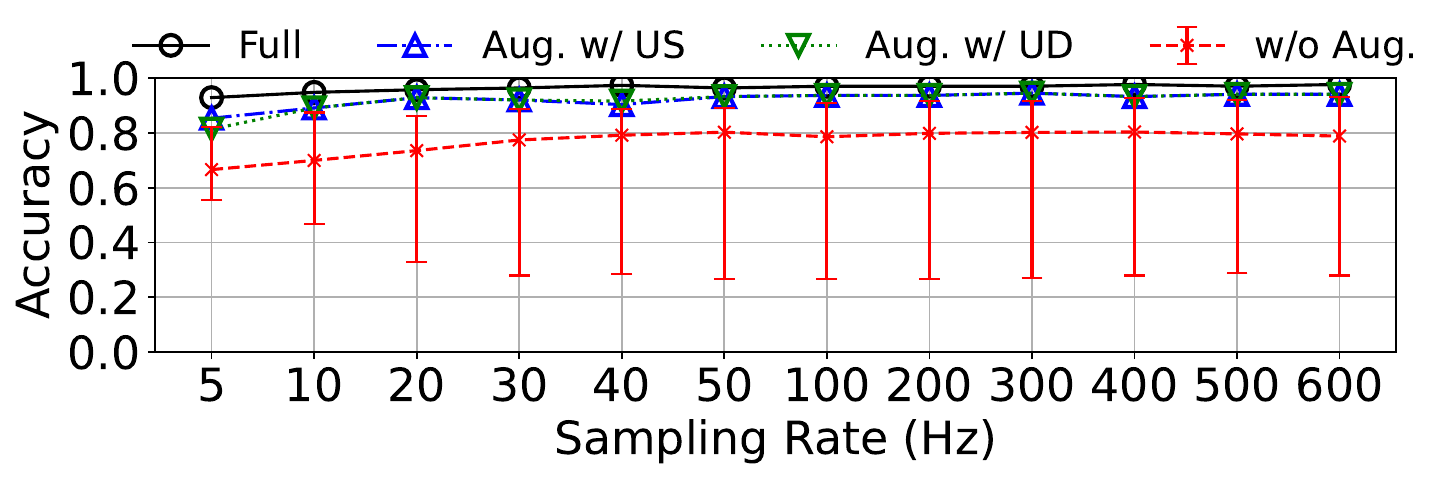}
  }\hfill
  \subfloat[]{
    \includegraphics[width=0.48\linewidth]{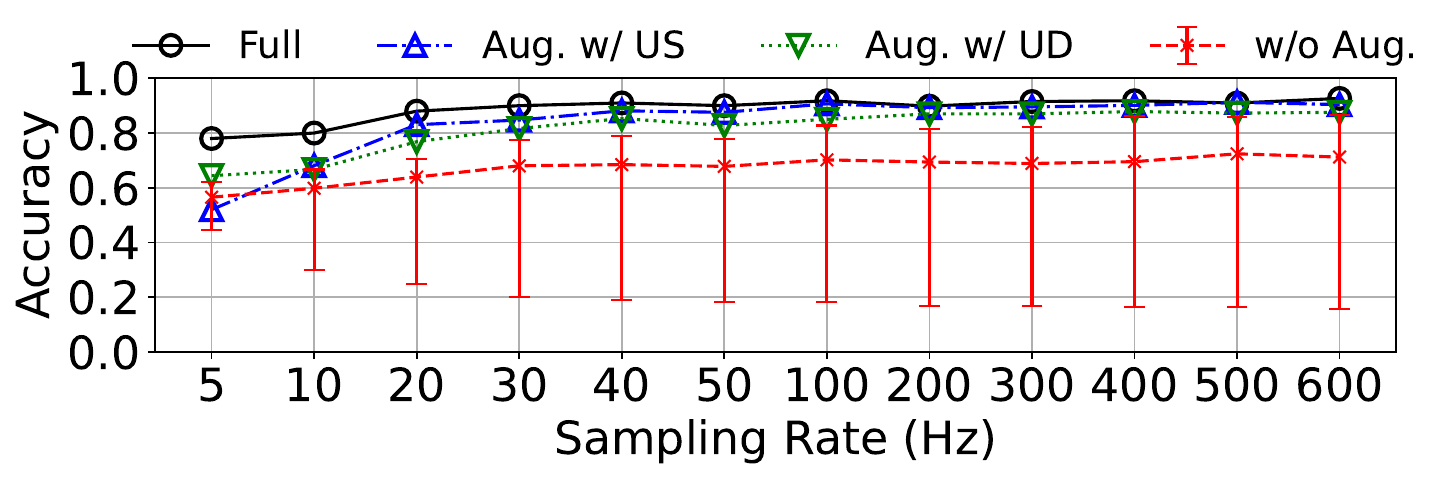}
  }

  \subfloat[]{
    \includegraphics[width=0.48\linewidth]{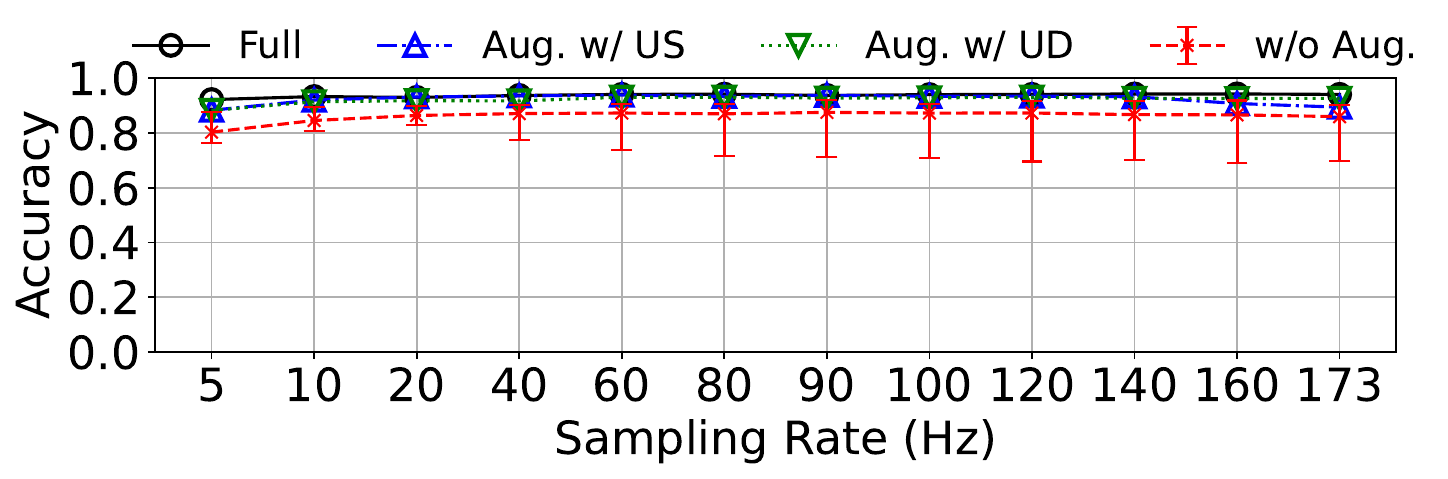}
  }\hfill
  \subfloat[]{
    \includegraphics[width=0.48\linewidth]{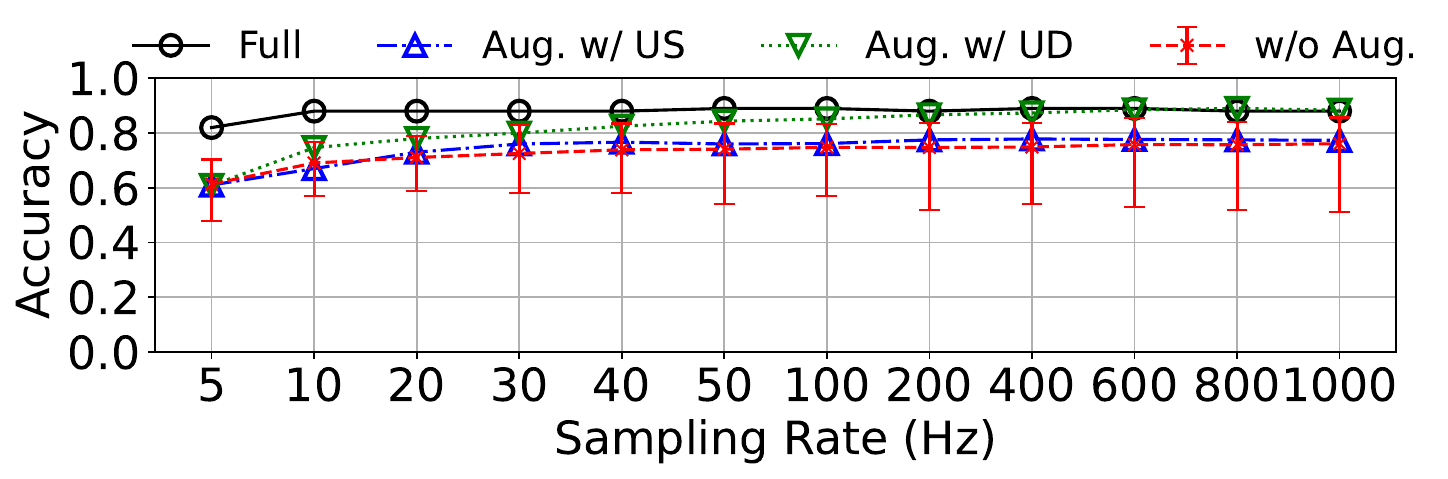}
  }
  \caption{Comparative performance of different models on different datasets with dynamic sampling rate augmentation. (a) SRV Activity dataset. (b) SRV Gesture dataset. (c) SHARP dataset. (d) Widar dataset.}
  \label{fig:lineplot}
\end{figure*}

\begin{table}[!t]
  \centering
  \caption{Comparative analysis of variance in accuracy across dynamic sampling rate augmentation with and without stochastic sampling and no augmentation.}
  \footnotesize %
  \begin{tabular}{@{}lcccc@{}}
  \toprule
  & Full & Aug. w/ US  &  Aug. w/ UD & w/o Aug.\\
  \midrule
  \textbf{SRV Activity}  & 0.00019 & 0.00133 & 0.00072 & 0.02820 \\
  \textbf{SRV Gesture}  & 0.00226 & 0.00667 & 0.01403 & 0.02693  \\
  \textbf{SHARP} & 0.00004 & 0.00019 & 0.00035 & 0.00294  \\
  \textbf{Widar} & 0.00036 & 0.00654 & 0.00273 & 0.00965  \\
  \bottomrule
\end{tabular}
  \label{tab:var_comparison}%
\end{table}%

We evaluate the full SRV-NN system trained with dynamic sampling rate augmentation, where the batch rate \(R_j\) is drawn from the adapted distribution and bounded between 5~Hz and each dataset's maximum sampling rate.

  The complete proposed system is denoted as ``Full'', while ``w/o Aug'' denotes the fixed-rate, conventionally trained baselines from Section~\ref{sec:srimpact}.

Fig.~\ref{fig:lineplot} and Table~\ref{tab:var_comparison} compare accuracy and accuracy variance, respectively. Because the 12 non-augmented SRV-NN models fluctuate substantially across test rates, error bars are used to show both the best achievable accuracy at each rate and the associated variation.

As shown in Fig.~\ref{fig:lineplot}, SRV-NN with the proposed augmentation consistently outperforms the conventionally trained baselines at every test sampling rate, with larger gains at lower rates. For example, on the SRV Activity dataset, comparing with the best baseline model, e.g., train-test on same sampling rate, our model improves accuracy by +8.3\% at 100 Hz (96.3\% vs 88.0\%) and +3.1\% at 600 Hz (96.7\% vs 93.6\%). Importantly, these results are achieved using a single model, whereas conventional training requires separate models for each sampling rate. This indicates that, the proposed method is able to generalise across different sampling rates without compromising the model's performance at a specific sampling rate.

In terms of averaged accuracy across the test sampling rates, the proposed model achieved average accuracies of 96.5\%, 88.8\%, 93.8\% and 87.8\% in SRV Activity, SRV Gesture, SHARP and Widar datasets, respectively. Those results are 19.5\%, 21.6\%, 12.1\% and 15.0\% higher than the averaged accuracy achieved by the baseline models.
In addition, as shown in Table~\ref{tab:var_comparison},
the SRV-NN trained with proposed augmentation shows lower accuracy variance across all four datasets. Taking the results in the SRV Activity dataset as an example, the accuracy variance for the augmentation-trained SRV-NN is $1.9 \times 10^{-4}$ which is significantly lower than that of baseline models ($2.8 \times 10^{-2}$).
\label{para:r2c4-results-summary}

Dynamic sampling rate augmentation therefore improves both accuracy and stability under changes in the test sampling rate. Packet loss and bursty traffic are common in real-world networks. They change the sensing input in two ways: they reduce the number of CSI sampling points in a sample, and they make the intervals between these points irregular. Because the test sets at each sampling rate are built with randomly selected sampling points, the rate sweep covers both effects at once. On SRV Activity, a 5~Hz test sample keeps only 0.8\% of the packets available at 600~Hz, which corresponds to a packet-loss ratio of 99.2\%, and the retained points arrive at random intervals. At 5~Hz, the packet-loss ratio is above 97\% on all four datasets. At this rate, the proposed system achieves 92.9\% accuracy on SRV Activity, compared with 55.1\% for a model trained only at 600~Hz.
\label{para:r2c4-robustness-analysis}

\subsubsection{Stochastic Sampling}\label{sec:ss_ablation}

Here, we evaluate the impact of stochastic sampling (SS) by comparing it with a uniform-sampling baseline (US), in which sampling points are selected at equal intervals during training. The performance of SRV-NN trained without stochastic sampling is shown by the blue dashed line in Fig.~\ref{fig:lineplot}. The corresponding accuracy variances across datasets are reported in Table~\ref{tab:var_comparison} under ``Aug. w/ US''.

As shown in Fig.~\ref{fig:lineplot}, stochastic sampling contributes most under low sampling rates. At 5~Hz, it improves accuracy by 11.5\%, 13.6\%, 4.0\%, and 21.2\% on SRV Activity, SRV Gesture, SHARP, and Widar, respectively. Randomising the sampling points during training gives the model more varied inputs at each sampling rate than uniform sampling, which reduces overfitting at low rates. It also exposes the model to the random intervals between real packet arrivals, discussed in Section~\ref{sec:motivation}.

As shown in Table~\ref{tab:var_comparison}, stochastic sampling also improves stability across test sampling rates. Models trained with stochastic sampling have a notably lower accuracy variance.
 For example, on SHARP, the model trained with stochastic sampling has a variance of $0.4 \times 10^{-4}$, much smaller than the $1.9 \times 10^{-4}$ of the model trained with uniform sampling.

The improvement becomes smaller at higher test sampling rates, with an average accuracy gain of 3.8\% at 100~Hz, because higher-rate inputs contain more detail and need less augmentation.

\subsubsection{Sampling Rate Distribution}
This subsection evaluates distribution adaptation using uniform distribution (UD), denoted as ``Aug. w/ UD'' in Fig.~\ref{fig:lineplot} and Table~\ref{tab:var_comparison}, as the baseline without adaptation.

The models that underwent training with the proposed distribution adaptation method significantly outperformed those trained without it. Our adaptive method results in an average accuracy improvement of 4.5\% and 5\% for the SRV Activity and SRV Gesture, respectively. For the SHARP and Widar datasets, it improved the accuracy by 1.5\% and 13.4\%, respectively.

In particular, distribution adaptation improves lower-rate performance most clearly, with the largest gain on Widar, where accuracy increases by 21.0\%. The gain decreases at higher rates, indicating that adaptation mainly strengthens low-rate generalisation.

Low sampling rates are inherently harder to generalise, as they carry less information and exhibit larger feature variation. By dynamically prioritising the rates on which the model underperforms, the proposed method trains broadly across the rate range while concentrating effort where generalisation is weakest, ultimately boosting the generalisability over variable sampling rates.

  \subsection{Ablation Studies}\label{sec:baseline_comparison}

\subsubsection{Augmentation Strategy Comparison}\label{sec:r1c6-r2c3-augmentation-strategy-comparison}To evaluate augmentation strategies under identical training and test conditions, we compare the proposed dynamic sampling rate augmentation with four benchmark methods in Table~\ref{tab:r2c3_classic_aug}. Three are classic time-series augmentations: Gaussian jittering~\cite{bishop1995training,an1996effects}, smooth time warping~\cite{um2017data} and magnitude scaling~\cite{um2017data}. We implement them with the public code from Iwana and Uchida~\cite{iwana2021empirical}. The fourth, trace-driven augmentation, is not a published method; we construct it for this benchmark to test whether measured Wi-Fi packet intervals can replace stochastic sampling. It uses 120-second packet-arrival traces of idle traffic, YouTube streaming, web browsing and email, because measured traffic preserves application-specific burstiness and idle gaps~\cite{paxson1995wide}, \cite{lan2002generation}. Each trace serves as a pool of candidate windows: for each training batch, the sampler selects one trace and takes a window as long as the CSI sample, e.g., a 2-s window for a 2-s sample, not the whole trace. The packet arrivals in this window are mapped to CSI temporal indices over the usable sequence length, then sorted and deduplicated.\par As shown in Table~\ref{tab:r2c3_classic_aug}, the proposed augmentation gives the highest average accuracy on all four datasets. It outperforms the strongest classic augmentation by 3.8\%, 16.4\%, 1.8\% and 8.1\% on SRV Activity, SRV Gesture, SHARP and Widar, respectively. Trace-driven augmentation also performs worse than the proposed method on all four datasets. Variable traffic changes the number of sampling points, but the classic augmentations do not: they change the recorded values or stretch and compress parts of the series. Trace-driven augmentation reproduces the gaps of the four measured traces, but it covers only the sampling rates and intervals that occur in these traces. The proposed augmentation varies both the sampling rate and the positions of the sampling points during training, and adaptively emphasises the sampling rates that are harder to generalise.

\begin{table}[!t]
  \centering
  \caption{Comparison of augmentation strategies under variable sampling rates. Avg. is the mean accuracy over the dataset-specific test sampling rates; Var. is scaled by $10^{-3}$. The best value in each column is highlighted in bold.}
  \setlength{\tabcolsep}{2pt}
  \renewcommand{\arraystretch}{0.9}
  \scriptsize
  \begingroup
  \resizebox{\columnwidth}{!}{\begin{tabular}{lcccccccc}
\toprule
& \multicolumn{2}{c}{SRV Activity}
& \multicolumn{2}{c}{SRV Gesture}
& \multicolumn{2}{c}{SHARP}
& \multicolumn{2}{c}{Widar} \bigstrut[t]\\
\cmidrule(lr){2-3}\cmidrule(lr){4-5}\cmidrule(lr){6-7}\cmidrule(lr){8-9}
Method & Avg. & Var. & Avg. & Var. & Avg. & Var. & Avg. & Var. \bigstrut[b]\\
\midrule
\textbf{Proposed} & \textbf{0.965} & 0.190 & \textbf{0.888} & \textbf{2.260} & \textbf{0.938} & \textbf{0.040} & \textbf{0.878} & \textbf{0.360} \bigstrut[t]\\
Trace-driven & 0.915 & \textbf{0.128} & 0.745 & 3.265 & 0.901 & 0.254 & 0.768 & 0.915 \\
Gaussian jittering & 0.927 & 0.320 & 0.713 & 7.289 & 0.919 & 0.104 & 0.797 & 4.062 \\
Smooth time warping & 0.911 & 0.238 & 0.724 & 6.583 & 0.920 & 0.177 & 0.787 & 3.710 \\
Magnitude scaling & 0.915 & 0.265 & 0.724 & 6.507 & 0.919 & 0.140 & 0.735 & 3.821 \bigstrut[b]\\
\bottomrule
\end{tabular} }
  \endgroup
  \label{tab:r2c3_classic_aug}
\end{table}
\label{para:r1c6-trace-driven-results}

\subsubsection[Backbone Ablation]{Backbone Ablation}
Section~\ref{sec:model_selection} compares the backbones without augmentation in order to select the backbone. Here, the same backbones are also trained with dynamic sampling rate augmentation, to test whether the gain of SRV-NN comes from the augmentation alone. Average accuracy and variance are reported over the same dataset-specific test sampling rates. Under fixed-rate training, each backbone is trained at 100~Hz, denoted ``\textit{model name} (100~Hz)'' in Table~\ref{tab:sota}, or at the maximum rate of each dataset, denoted ``\textit{model name} (Max rate)''. The maximum rates are 600~Hz for SRV Activity and SRV Gesture, 173~Hz for SHARP, and 1000~Hz for Widar. Each test sample is then resampled to the training rate of the model before recognition by linear interpolation, as used in~\cite{falldefi} to fill lost packets. When trained with augmentation, denoted ``\textit{model name} w/ Aug.'', CNN and CNN-GRU receive their training and test sequences after linear interpolation to the length of a 100~Hz sequence. This is needed because their convolutional layers require a minimum input length (Section~\ref{sec:model_selection}). GRU and BiLSTM take the variable-length sequences directly and classify from the final hidden state.
\label{para:r1c5-aug-baseline-protocol}

  \begin{table}[!t]
    \centering
    \caption{Baseline comparison with fixed-rate training and dynamic sampling rate augmentation. The reported variance (Var.) values are scaled by $10^{-3}$. The best value in each column is highlighted in bold.}

    \setlength{\tabcolsep}{2pt}%
    \renewcommand{\arraystretch}{0.9}%
    \scriptsize
    \resizebox{\columnwidth}{!}{%
      \begin{tabular}{lcccccccc}
\toprule
& \multicolumn{2}{c}{SRV Gesture}
& \multicolumn{2}{c}{SRV Activity}
& \multicolumn{2}{c}{SHARP}
& \multicolumn{2}{c}{Widar} \bigstrut[t]\\
\cmidrule(lr){2-3}\cmidrule(lr){4-5}\cmidrule(lr){6-7}\cmidrule(lr){8-9}
& Avg. & Var. & Avg. & Var. & Avg. & Var. & Avg. & Var. \bigstrut[b]\\
\midrule
CNN (100~Hz)        & 0.770 & 5.200  & 0.850 & 1.800  & 0.830 & 5.480  & 0.330 & 23.500 \bigstrut[t]\\
CNN (Max rate)  & 0.670 & 9.300  & 0.790 & 16.100 & 0.700 & 33.200 & 0.400 & 35.800 \\
CNN w/ Aug.        & 0.806 & 15.783 & 0.913 & 0.447  & 0.878 & 1.219  & 0.700 & 13.262 \\
\midrule
CNN-GRU (100~Hz)    & 0.520 & 3.700  & 0.860 & 0.400  & 0.520 & 26.800 & 0.410 & 24.700 \\
CNN-GRU (Max rate) & 0.410 & 4.760  & 0.910 & 1.180  & 0.600 & 51.100 & 0.390 & 42.500 \\
CNN-GRU w/ Aug.    & 0.560 & 3.157  & 0.913 & \textbf{0.151}  & 0.780 & 0.854  & 0.502 & 3.068 \\
\midrule
BiLSTM (100~Hz)     & 0.610 & 3.900  & 0.890 & 0.180  & 0.550 & 32.000 & 0.230 & 49.900 \\
BiLSTM (Max rate) & 0.550 & 9.630  & 0.910 & 0.179  & 0.610 & 35.000 & 0.490 & 11.300 \\
BiLSTM w/ Aug.     & 0.669 & 11.960 & 0.911 & 0.330  & 0.803 & 2.992  & 0.633 & 10.026 \\
\midrule
GRU (100~Hz)        & 0.610 & 28.000 & 0.870 & 1.150  & 0.620 & 46.000 & 0.400 & 20.800 \\
GRU (Max rate)  & 0.680 & 7.620  & 0.850 & 2.890  & 0.520 & 59.200 & 0.360 & 53.200 \\
GRU w/ Aug.        & 0.682 & 2.460  & 0.885 & 1.736  & 0.696 & 8.016  & 0.499 & 2.667 \\
\midrule
SRV-NN w/ Aug.      & \textbf{0.890} & \textbf{2.100} & \textbf{0.960} & 0.170 & \textbf{0.930} & \textbf{0.052} & \textbf{0.870} & \textbf{0.354} \bigstrut[b]\\
\bottomrule
\end{tabular} %
    }
    \label{tab:sota}%
  \end{table}%

  Table~\ref{tab:sota} presents comparison results across all four datasets.
With augmentation, every baseline model reaches a higher average accuracy on all four datasets, and most baseline configurations show a lower variance. The augmentation strategy therefore contributes to robustness. SRV-NN nevertheless remains the strongest model: it achieves the highest average accuracy on all four datasets and the lowest variance on SRV Gesture, SHARP and Widar. On SRV Activity, CNN-GRU w/ Aug. has a slightly lower variance, but its average accuracy is lower than that of SRV-NN. SRV-NN is therefore the most effective model for generalising across sampling rates.
\label{para:r1c5-aug-baseline-results}

\subsubsection[SRV-NN Classifier Ablation]{SRV-NN Classifier Ablation}

  \begin{table}[!t]
    \centering
    \caption{SRV-NN classifier ablation under dynamic sampling rate augmentation. The reported variance (Var.) values are scaled by $10^{-3}$. The best value in each column is highlighted in bold.}
    \setlength{\tabcolsep}{2pt}
    \renewcommand{\arraystretch}{0.9}
    \scriptsize
    \begingroup
    \resizebox{\columnwidth}{!}{\begin{tabular}{lcccccccc}
\toprule
& \multicolumn{2}{c}{SRV Gesture}
& \multicolumn{2}{c}{SRV Activity}
& \multicolumn{2}{c}{SHARP}
& \multicolumn{2}{c}{Widar} \bigstrut[t]\\
\cmidrule(lr){2-3}\cmidrule(lr){4-5}\cmidrule(lr){6-7}\cmidrule(lr){8-9}
& Avg. & Var. & Avg. & Var. & Avg. & Var. & Avg. & Var. \bigstrut[b]\\
\midrule
SRV-NN w/ max pooling (Ours) & \textbf{0.890} & \textbf{2.100} & \textbf{0.960} & 0.170 & \textbf{0.930} & \textbf{0.052} & \textbf{0.870} & \textbf{0.354} \bigstrut[t]\\
SRV-NN w/ average pooling   & 0.726 & 7.150 & 0.915 & \textbf{0.068} & 0.907 & 0.204 & 0.753 & 6.489 \\
SRV-NN w/ CLS token         & 0.749 & 7.644 & 0.940 & 0.132 & 0.917 & 0.303 & 0.631 & 5.313 \\
SRV-NN w/ attention pooling & 0.717 & 6.240 & 0.922 & 0.125 & 0.894 & 0.249 & 0.738 & 7.604 \bigstrut[b]\\
\bottomrule
\end{tabular} }
    \endgroup
    \label{tab:r1c5_readout_ablation}%
  \end{table}%
To evaluate the classifier design alone, we keep the transformer backbone and the augmentation of SRV-NN fixed and replace the sampling rate versatile classifier with three benchmark classifiers.
The average-pooling classifier computes the mean of all feature vectors~\cite{lee2021learnable}.
The CLS-token classifier prepends a learnable vector to the input sequence and classifies its final transformer output~\cite{devlin2019bert}.
The attention-pooling classifier learns a score for each feature vector, normalises the scores over all feature vectors with a softmax, and forms a weighted sum of the feature vectors~\cite{lin2017structured}.
Each of these classifiers gives the feature vectors weights that sum to one over the whole sequence: equal weights for average pooling and softmax-normalised attention weights for the other two. Their outputs therefore change when a different sampling rate adds or removes feature vectors.
Table~\ref{tab:r1c5_readout_ablation} shows that max pooling achieves the highest average accuracy on all four datasets and the lowest variance on SRV Gesture, SHARP and Widar. The three benchmark classifiers have a lower variance on SRV Activity, but also a lower average accuracy, so max pooling gives the best overall trade-off between accuracy and stability.
\label{para:r1c5-readout-definitions}

\subsection[Evaluation under Measured Packet-Arrival Traces]{Evaluation under Measured Packet-Arrival Traces}\label{sec:r1c2-traffic-eval}
To evaluate the proposed method under real traffic, we use the four measured packet-arrival traces of Section~\ref{sec:r1c6-r2c3-augmentation-strategy-comparison} to decide which CSI sampling points of each test sample are kept. This is similar to the trace-driven augmentation, but the traces are used at a different stage. The trace-driven augmentation applies them to training batches; but here, it is applied to test samples, to simulate real communication traffic.
For a test sample of duration $T$ (Table~\ref{tab:dataset}), we take a $T$-second window at a random position in a trace. We then keep only the CSI sampling points that correspond to the packet arrivals in this window. Windows with an average sampling rate below 5~Hz, the lowest test sampling rate, are not used. We repeat the test ten times on each dataset and trace with different random windows. Table~\ref{tab:r1c2_traffic_eval} reports the average accuracy, where ``Reference'' denotes the same test samples with all CSI sampling points kept.

\begin{table}[t]
\centering
\caption{Recognition accuracy under measured traffic traces. The reference column retains every CSI observation in the recording, without traffic-driven removal.}
\label{tab:r1c2_traffic_eval}
\setlength{\tabcolsep}{3pt}
\begingroup
\resizebox{\columnwidth}{!}{\begin{tabular}{lccccc}
\toprule
\textbf{Dataset} & \textbf{Idle} & \textbf{YouTube} & \textbf{Web} & \textbf{Email} & \textbf{Reference} \bigstrut[t]\\
\midrule
SRV Activity & 0.9099 & 0.9425 & 0.9077 & 0.9369 & 0.9776 \bigstrut[t]\\
SRV Gesture  & 0.8136 & 0.8678 & 0.8223 & 0.8534 & 0.8983 \\
SHARP        & 0.9175 & 0.9197 & 0.9156 & 0.9238 & 0.9357 \\
Widar        & 0.8280 & 0.8707 & 0.8650 & 0.8172 & 0.8983 \bigstrut[b]\\
\bottomrule
\end{tabular} }
\endgroup
\end{table}

Under all four measured traces, the accuracy of the proposed method stays above 0.90 on SRV Activity and SHARP and above 0.81 on SRV Gesture and Widar. Because these traces include low packet rates and bursty, irregular arrivals, the results show that the proposed method remains robust under real traffic.
\label{para:r1c2-traffic-eval-results}

  \subsection{Sensitivity Analysis}
  \subsubsection{Distribution Adaptation Learning Rate}
  To evaluate the impact of the distribution adaptation learning rate $\alpha$, we systematically vary it from 0.1 to 4.0 across all datasets and assess each configuration using the dual-metric evaluation framework of Section~\ref{sec:eval_metrics}, i.e., average accuracy and variance across different sampling rates.

  As Table~\ref{tab:sensi} shows, the optimal $\alpha$ appears to be task-specific: for activity recognition tasks (SRV Activity and SHARP), $\alpha$ in the range [0.3, 0.9] maintains a consistently high average accuracy ($\geq 0.91$) with low variance, whereas gesture recognition tasks (SRV Gesture and Widar) are more sensitive and degrade significantly once $\alpha$ exceeds 1.0; on Widar, the average accuracy drops from 0.87 at $\alpha$ = 0.7 to 0.25 at $\alpha$ = 3.5. We therefore empirically selected $\alpha$ = 0.7, which balances average accuracy against variance across all datasets and generalises robustly without task-specific tuning.

\begin{table}[!t]
  \centering
  \caption{Impacts of the distribution adaptation learning rate $\alpha$.
  The reported variance (Var.) values are scaled by $10^{-3}$.}

  \setlength{\tabcolsep}{7pt}
  \renewcommand{\arraystretch}{0.9}%
  \scriptsize
  \resizebox{\columnwidth}{!}{%
  \begin{tabular}{ccccccccc}
    \toprule
    & \multicolumn{2}{c}{SRV Gesture}
    & \multicolumn{2}{c}{SRV Activity}
    & \multicolumn{2}{c}{SHARP}
    & \multicolumn{2}{c}{Widar} \bigstrut[t]\\
    \cmidrule(lr){2-3}\cmidrule(lr){4-5}\cmidrule(lr){6-7}\cmidrule(lr){8-9}
    \textbf{$\alpha$}
    & Avg. & Var.
    & Avg. & Var.
    & Avg. & Var.
    & Avg. & Var. \bigstrut[b]\\
    \midrule
    \textbf{0.1} & 0.72 & 2.28  & 0.91 & 0.32 & 0.92  & 0.14  & 0.76 & 0.52 \bigstrut[t]\\
    \textbf{0.3} & 0.68 & 5.76  & \textbf{0.92} & 0.31 & 0.92  & 0.184  & 0.81 & 0.58 \\
    \textbf{0.5} & 0.84 & 7.41  & \textbf{0.92} & 0.34 & 0.92  & 0.133  & 0.85 & 0.39 \\
    \textbf{0.7} & \textbf{0.89} & \textbf{2.34}  & 0.91 & \textbf{0.20} & \textbf{0.93} & \textbf{0.02} & \textbf{0.87} & 0.35 \\
    \textbf{0.9} & 0.87 & 6.09  & 0.92 & 0.23 & 0.93 & 0.04 & 0.81 & 0.35 \\
    \textbf{1.1} & 0.66 & 4.00  & 0.94 & 0.45 & 0.93 & 0.08 & 0.72 & 2.82  \\
    \textbf{1.3} & 0.70 & 3.97  & 0.92 & 0.15 & 0.92 & 0.08 & 0.56 & 4.93  \\
    \textbf{1.5} & 0.69 & 6.47  & 0.92 & 0.27 & 0.93 & 0.06 & 0.76 & 0.84 \\
    \textbf{2.0} & 0.73 & 8.77  & 0.78 & 2.35  & 0.92 & 0.18  & 0.66 & 4.33  \\
    \textbf{2.5} & 0.72 & 5.13  & 0.91 & 0.26 & 0.85 & 0.95 & 0.58 & 4.48  \\
    \textbf{3.0} & 0.72 & 6.67 & 0.91 & 0.22 & 0.94 & 0.04 & 0.61 & 4.89  \\
    \textbf{3.5} & 0.65 & 13.1 & 0.91 & 0.18 & 0.89 & 0.86 & 0.25 & 3.13  \\
    \textbf{4.0} & 0.66 & 8.66 & 0.92 & 0.32 & 0.89 & 0.90 & 0.32 & \textbf{0.14} \bigstrut[b]\\
    \bottomrule
  \end{tabular}%
  }
  \label{tab:sensi}%
\end{table}%
\subsubsection{Learning Rate}

\begin{figure}
  \centering
  \includegraphics[width=1\linewidth]{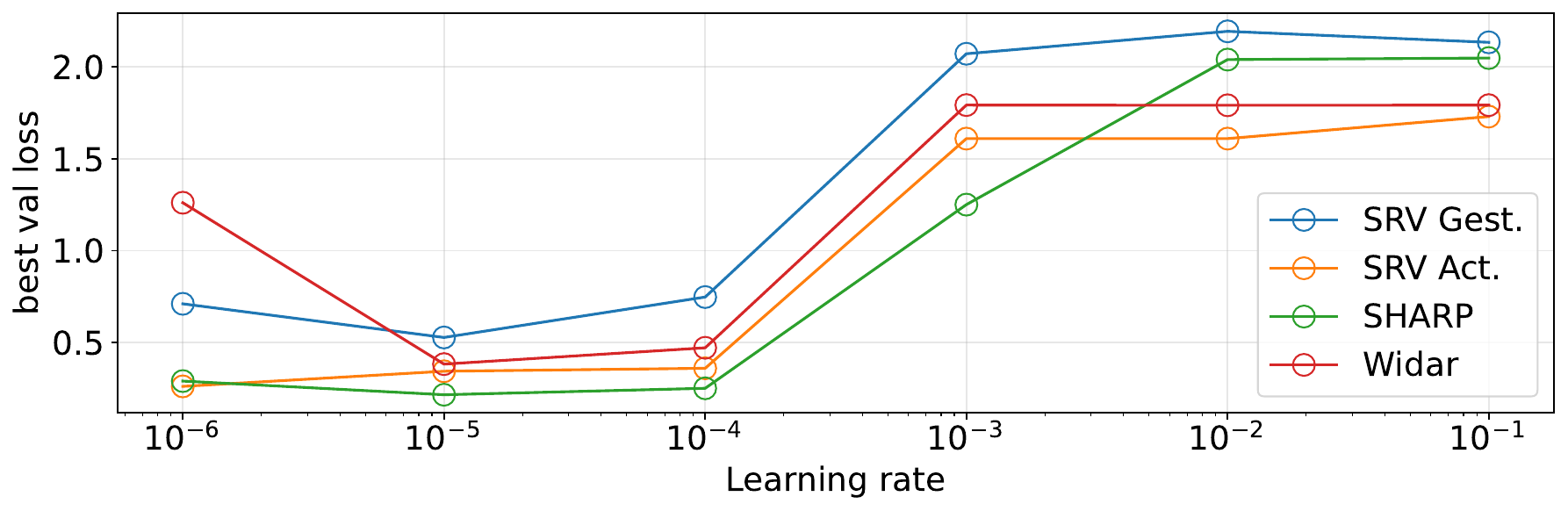}
  \caption{Learning rate sensitivity over all datasets}
  \label{fig:lrsensitivity}
\end{figure}

To evaluate the robustness of SRV-NN to the learning rate hyperparameter, we now conduct experiments across a range of learning rates from $10^{-6}$ to $10^{-1}$. The validation loss curves across all four datasets exhibit consistent patterns, as presented in Fig.~\ref{fig:lrsensitivity}.

The model generally performs best at learning rates from $10^{-5}$ to $10^{-4}$, where the validation loss remains consistently low across all datasets and SRV-NN shows stable training dynamics and robust convergence. At high learning rates ($10^{-3}$ and $10^{-1}$), however, the validation loss increases substantially on all datasets, indicating training instability and poor convergence caused by excessively large parameter updates. SRV-NN therefore remains robust across learning rates from $10^{-6}$ to $10^{-4}$, i.e., it is not overly sensitive to this hyperparameter within this range, and we adopted $10^{-5}$ for all experiments reported in this paper.

  \subsection{Computational Complexity Analysis}\label{sec:computational_complexity}
  \begin{figure}
    \centering
    \includegraphics[width=0.99\linewidth]{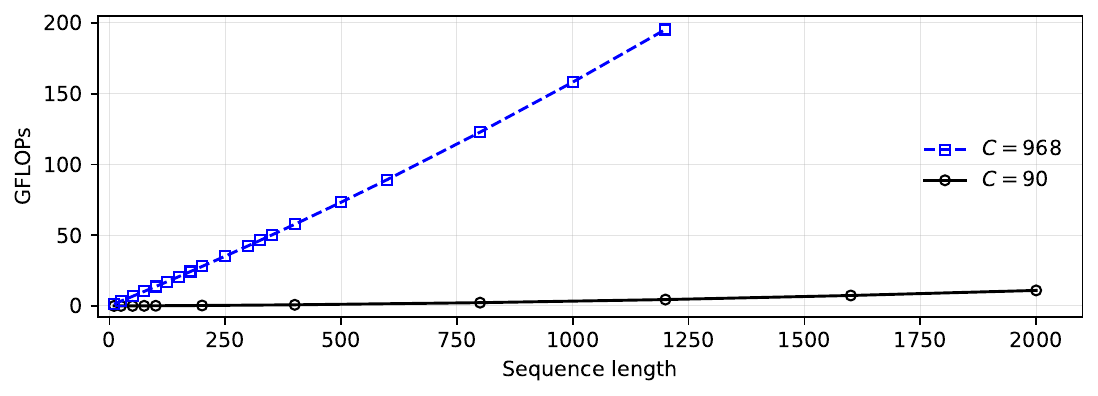}
    \caption{Transformer inference cost versus sequence length for the two evaluated CSI feature dimensions.}
    \label{fig:computational_complexity}
  \end{figure}

\label{para:r3c2-compute-protocol}
For practical deployment, we profile the model size and the number of multiply--accumulate operations (MACs). The cost depends on the CSI feature dimension $C$ and the sequence length $N$. For each of the two feature dimensions, we evaluate the longest input in our datasets: $(C,N)=(968,1200)$ for SRV Activity at 600~Hz and $(C,N)=(90,2000)$ for Widar at 1000~Hz. The learned positional encoding of SRV-NN is sized for this longest $N$. The parameter counts and GFLOPs in Table~\ref{tab:computational_comparison} are computed for each model under these two input configurations. GFLOPs count two floating-point operations per MAC and include matrix, convolution, recurrent and attention MACs. Non-MAC operations such as softmax, normalisation, activation, pooling and masking are excluded.

\label{para:r3c2-compute-discussion}
Fig.~\ref{fig:computational_complexity} shows how the operation count of SRV-NN grows with $N$ for the two values of $C$. The growth is steeper for $C=968$ because of the higher-dimensional input. Table~\ref{tab:computational_comparison} compares SRV-NN with all baselines at these two configurations in terms of parameter count and GFLOPs.

\begin{table}[!t]
  \centering
  \caption{Model size and GFLOPs for the longest input of each CSI feature dimension.}
  \label{tab:computational_comparison}
  \begingroup
  \setlength{\tabcolsep}{2.2pt}
  \resizebox{\columnwidth}{!}{\begin{tabular}{lrrrr}
\toprule
& \multicolumn{2}{c}{$(C,N)=(968,1200)$} & \multicolumn{2}{c}{$(C,N)=(90,2000)$} \\
\cmidrule(lr){2-3}\cmidrule(lr){4-5}
Model & Params (M) & GFLOPs & Params (M) & GFLOPs \\
\midrule
CNN & 4.103 & 204.560 & 4.031 & 30.642 \\
CNN-GRU & 2.195 & 1.001 & 2.083 & 1.222 \\
GRU & 2.279 & 5.456 & 0.931 & 3.699 \\
BiLSTM & 24.974 & 59.847 & 21.379 & 85.361 \\
SRV-NN & 68.684 & 195.372 & 0.769 & 10.973 \\
\bottomrule
\end{tabular} }
  \endgroup
\end{table}

\label{para:r3c2-compute-profile}

\label{para:r2c5-training-overhead}
Dynamic sampling rate augmentation adds little computation. It only downsamples the CSI before the forward pass, and it adds no model layers and no operations at inference. Because the downsampled sequences are shorter than the original high-rate sequences, it can also reduce the training computation.

\section{Conclusion}\label{sec:conclusion}
This paper presented SRV-NN and dynamic sampling rate augmentation for Wi-Fi sensing under variable sampling rates. SRV-NN combines a transformer feature extractor with a sampling rate versatile classifier to address variable input length, while dynamic sampling rate augmentation improves robustness to variable rates and sampling intervals. Evaluated on SRV Activity, SRV Gesture, SHARP, and Widar, the augmented SRV-NN achieved average accuracies of 96.5\%, 88.8\%, 93.8\%, and 87.8\%, with variances of $0.19 \times 10^{-3}$, $2.26 \times 10^{-3}$, $0.04 \times 10^{-3}$, and $0.36 \times 10^{-3}$, respectively.
Overall, the proposed system demonstrates strong sampling rate generalisation for real-world scenarios with variable packet arrivals.
A limitation of the proposed system is that its accuracy is bounded by the motion information that the received CSI sequence contains. Under extreme packet loss or highly bursty traffic, a fast or subtle motion may fall entirely within a long gap between CSI sampling points. The proposed augmentation improves robustness to changes in sampling rate and sampling interval, but it cannot recover motion details that are absent from the input CSI.
\label{para:r2c4-r2c5-limitations}

\bibliographystyle{IEEEtran}
\bibliography{IEEEabrv,mybibfile}
\end{document}